\documentclass{article}

\usepackage[final]{corl_2022} 

\usepackage{graphics} 
\usepackage{epsfig} 
\usepackage{amsmath} 
\usepackage{amssymb}  

\usepackage[utf8]{inputenc} 
\usepackage[T1]{fontenc}    

\usepackage[dvipsnames]{xcolor}
\usepackage{url}            
\usepackage{booktabs}       
\usepackage{amsfonts}       
\usepackage{nicefrac}       
\usepackage{microtype}      

\usepackage{subfigure}
\usepackage{booktabs} 
\usepackage{multirow}
\usepackage{pifont}
\usepackage[noend]{algpseudocode}
\usepackage{setspace}
\usepackage{comment}
\usepackage[ruled]{algorithm2e}
\usepackage{verbatim}
\usepackage{caption}
\usepackage{wrapfig}
\usepackage{marvosym}

\usepackage{hyperref}
\hypersetup{
    colorlinks=true,
    filecolor=magenta,      
}

\definecolor{revisioncolor}{rgb}{0.0, 0.0, 0.0}

\newcommand{\OODIL}{{OOD-IL}}

\title{Out-of-Dynamics Imitation Learning from Multimodal Demonstrations}

\author{
  Yiwen Qiu$^1$, Jialong Wu$^2$, Zhangjie Cao$^3$, Mingsheng Long$^2$\textsuperscript{\Letter}\\
  $^1$Department of Automation, Tsinghua University, China\\
  $^2$School of Software, BNRist, Tsinghua University, China \\
  $^3$Department of Computer Science, Stanford University, Stanford, CA 94305, USA \\
  \texttt{\{qywmei,wujialong0229\}@gmail.com}\\
  \texttt{caozj@cs.stanford.edu, mingsheng@tsinghua.edu.cn}
}

\begin{document}

\maketitle

\begin{abstract}
Existing imitation learning works mainly assume that the demonstrator who collects demonstrations shares the same dynamics as the imitator. However, the assumption limits the usage of imitation learning, especially when collecting demonstrations for the imitator is difficult. In this paper, we study out-of-dynamics imitation learning (OOD-IL), which relaxes the assumption to that the demonstrator and the imitator have the same state spaces but could have different action spaces and dynamics. OOD-IL enables imitation learning to utilize demonstrations from a wide range of demonstrators but introduces a new challenge: some demonstrations cannot be achieved by the imitator due to the different dynamics. 
Prior works try to filter out such demonstrations by feasibility measurements, but ignore the fact that the demonstrations exhibit a multimodal distribution since the different demonstrators may take different policies in different dynamics, which hinders learning an accurate measurement. 
We develop a better transferability measurement to tackle this newly-emerged challenge. We first design a novel sequence-based contrastive clustering algorithm to cluster demonstrations from the same mode to avoid the mutual interference of demonstrations from different modes and then learn the transferability of each demonstration with an adversarial-learning based algorithm in each cluster. Experiment results on several MuJoCo environments, a driving environment and a simulated robot environment show that the proposed transferability measurement more accurately finds and down-weights non-transferable demonstrations and outperforms prior works on the final imitation learning performance. We show the videos of our experiment results on our \href{https://sites.google.com/view/oodil}{\color{mydarkblue} website}.
\end{abstract}

\keywords{Imitation Learning, Out-of-Dynamics Imitation Learning}

\section{Introduction}

Imitation learning is a widely-used policy learning paradigm to solve robotics and control tasks, which learns the policy from demonstrations~\cite{zhang2018deep,codevilla2018end}. Standard imitation learning assumes that the demonstrator who collects the demonstrations shares the same dynamics with the imitator, which means that their state spaces, action spaces and the transition models are all the same~\cite{ross2011reduction,ziebart2008maximum,ho2016generative}. However, such strict assumption limits the practical usage of imitation learning, especially when it is difficult to collect demonstrations in the imitator's environment.

In this paper, we relax the assumption to that the demonstrator and the imitator share the same state space but the dynamics could be different, i.e., their action spaces and the transition models could be different. We name the new imitation learning setting as out-of-dynamics imitation learning (\OODIL). The new assumption poses fewer requirements on the demonstrator and enables imitation learning to utilize a broader range of demonstrations.
For example, in autonomous driving, to learn the policy of an autonomous vehicle in a new city, we may use the database of driving behavior of human vehicles containing useful information instead of manually collecting demonstrations on the autonomous vehicle. But such a database cannot be used by standard imitation learning since the vehicles have different dynamics and the city environments can be different. Furthermore, in real applications, it is often difficult and expensive to obtain enough data from a single source, thus it is beneficial to utilize 
data from a mixture of massive data sources for better real-world performance.

\begin{figure}[t]
    \centering
    \subfigure[Problem setting and the general framework.]{\includegraphics[width=.71\textwidth]{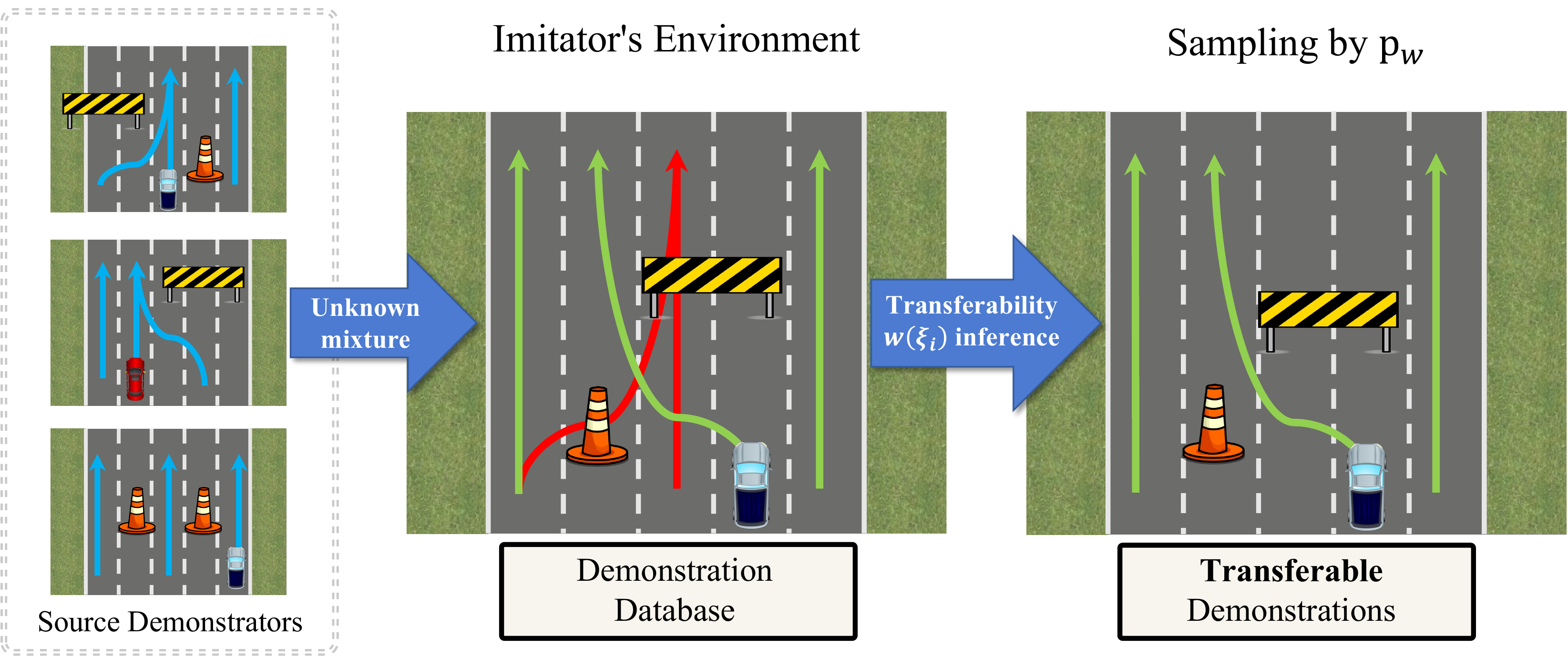}\label{fig:front_fig}}
    \subfigure[Multimodal distribution.]{\includegraphics[width=0.28\textwidth]{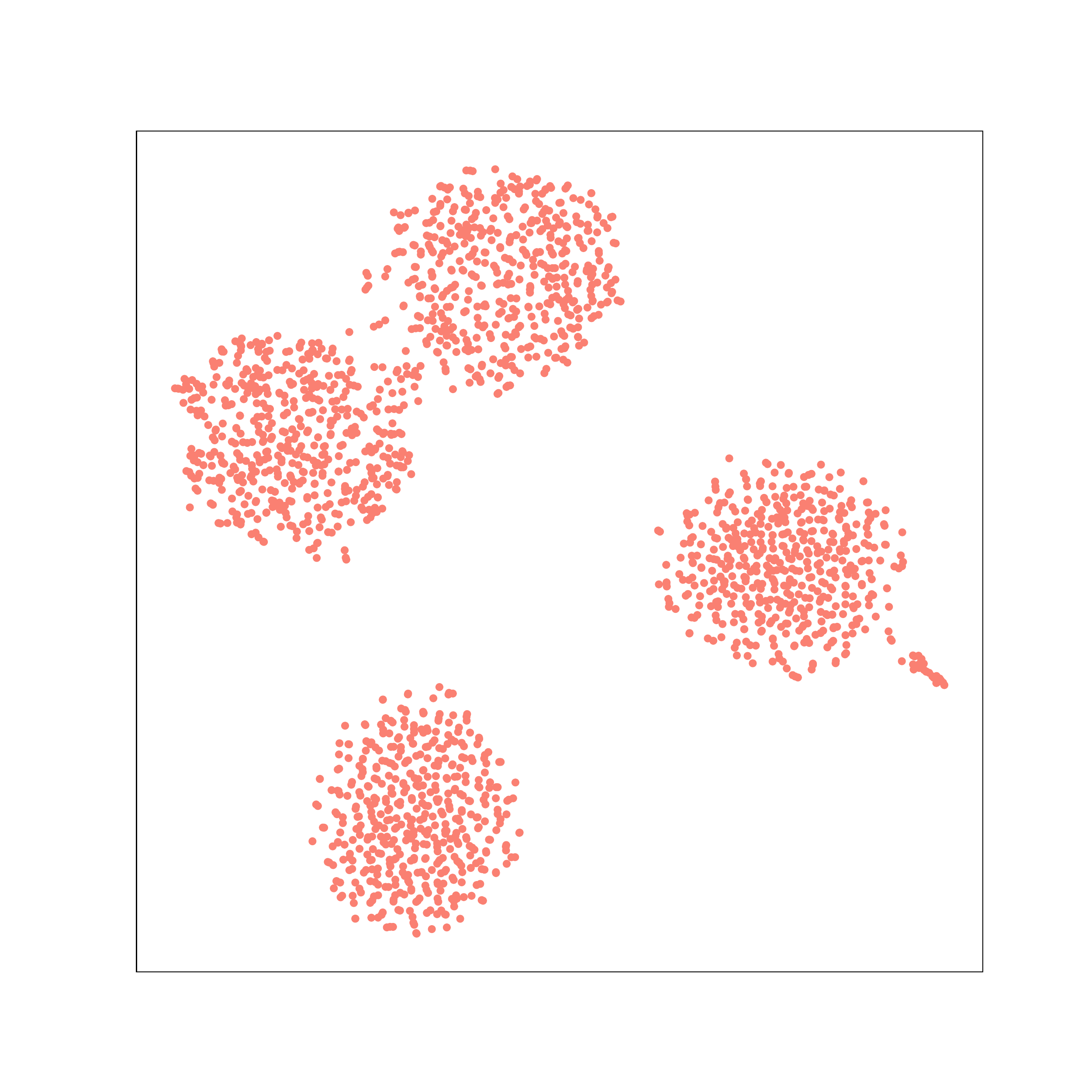}\label{fig:multimodal}}
    \vspace{-3pt}
    \caption{(a) We are given demonstrations collected from an unknown mixture of demonstrators, which exhibit a multimodal distribution since different demonstrators may take different policies. Some demonstrations (red) may not be achievable by the target agent while others (green) are achievable. Learning from these unachievable demonstrations introduces unknown behaviors. We develop a transferability measurement to mitigate negative impact of these non-transferable demonstrations and learn from transferable ones. (b) A t-SNE visualization of demonstrations collected from $4$ dynamics in the MuJoCo environment. }
    \vspace{-12pt}
\end{figure}

As shown in Figure~\ref{fig:front_fig}, \OODIL\ introduces a challenge that the mixture of demonstrations collected in different dynamics may not be achievable by the imitator, and thus are non-transferable. The state-of-the-art work constructs feasibility measurement to down-weight the non-transferable demonstrations by learning a unimodal policy from a feasibility-MDP (f-MDP)~\citep{cao2021corl}. However, as shown in Figure~\ref{fig:multimodal}, we notice that the prior work overlooks an important fact: the demonstrations collected from different demonstrators exhibit a multimodal distribution since different demonstrators may take different policies. The multimodal demonstrations make it difficult to learn a unimodal policy from the f-MDP as well as learn an accurate feasibility measurement. Furthermore, f-MDP suffers from slow and inaccurate optimization difficulties due to a step-by-step optimization procedure.

In this paper, we address \OODIL\ with the above challenges by developing a better transferability measurement to determine how transferable each demonstration is for the imitator. \textit{To fully remove the interference from multimodal distribution}, we design a sequence-based contrastive clustering algorithm to simultaneously learn a hidden space and cluster the trajectories in the hidden space, which ensures each cluster introduces a unimodal trajectory distribution. Then we learn the transferability for each cluster respectively by generative adversarial imitation learning (GAIL)~\cite{ho2016generative}, a much easier-to-optimize objective than f-MDP. The discriminator in GAIL distinguishes transitions from the demonstrators' distribution to the imitators' distribution, which serves as the transferability measurement to indicate the likelihood for a demonstration trajectory to be reproduced by the imitator. Experiment results on several MuJoCo environments, a driving environment, and a simulated Franka Panda Arm environment show that by reweighting demonstrations with the proposed transferability measurement, the final imitation learning policy outperforms all baselines and achieves the state-of-the-art performance.

\section{Related Works}\label{sec:related_work}

\noindent \textbf{Standard Imitation Learning.} Imitation learning learns a policy to imitate the behavior in demonstrations. Existing algorithms can be roughly categorized into three types: Behavior Cloning (BC), Inverse Reinforcement Learning (IRL), and Generative Adversarial Imitation Learning (GAIL).
BC utilizes supervised learning to directly learn the policy from state-action pairs~\cite{bain1995framework}. Following works propose dataset aggregation~\cite{ross2011reduction} or policy aggregation~\cite{daume2009search,ross2010efficient} to address the compounding errors problem in BC. Torabi \textit{et al.} \cite{torabi2018behavioral} try to learn from state sequences by first recovering the actions through an inverse dynamics model and then conducting behavior cloning.
IRL first recovers the reward function from demonstrations and then learns the policy using the learned reward~\cite{abbeel2004apprenticeship,ng2000algorithms,ziebart2008maximum,finn2016guided}. 
GAIL-based works match the occupancy measure between the learned policy and the demonstrations through adversarial learning to seek the optimal policy~\cite{ho2016generative,fu2017learning}. Also, GAIL is demonstrated to successfully imitate state sequences~\cite{schroecker2017state,torabi2018generative,sun2019provably}.
However, all the methods assume that the demonstrator and the imitator share the same dynamics, which violates the assumption of \OODIL.

\noindent \textbf{Out-of-dynamics Imitation Learning.} 
To address the \OODIL\ problem, there are works learning a correspondence model~\cite{nehaniv2002correspondence} between the demonstrator and the imitator~\cite{englertaddressing,calinon2007learning,eppner2009imitation,zhang2021learning,wang2022weakly}. However, these methods assume that strict correspondence, i.e. a one-to-one mapping between the state spaces and action spaces, exists between the demonstrator and the imitator. This assumption can be violated in real-world applications, e.g., a 7-DoF robot arm and a 3-DoF robot arm have no such correspondence since some behavior of the 7-DoF robot cannot be realized by the 3-DoF robot. Recent works relax the assumption to that the demonstrator and the imitator only share the state space~\cite{liu2019state,cao2021learning}. One line of work aims to maximally follow the demonstrations~\cite{liu2019state,sharma2018multiple,Gangwani2020State-only,radosavovic2020state}, but following the non-transferable demonstrations is impossible. 

Cao \textit{et al.}~\cite{cao2021learning} develop a feasibility measurement to down-weight non-transferable demonstrations by learning an inverse dynamics model, but the learned inverse dynamics may not generalize to all the demonstrations. To address this difficulty, the state-of-the-art work improves the feasibility by learning the optimal policy in f-MDP~\cite{cao2021corl}. However, f-MDP suffers from two limitations. Firstly, f-MDP enforces one unimodal policy to maximally imitate the demonstrations but the demonstrations with a multimodal distribution are hard to be modeled by such a policy.
Secondly, at each time step, only when the policy is optimized in all prior time steps can it be optimized to maximize the reward at the current time step. Such inappropriate design makes it difficult and inefficient to learn the optimal policy of f-MDP. We instead use the discriminator in GAIL to learn the transferability measurement.

\textcolor{revisioncolor}{
\noindent \textbf{Contrastive Clustering.} Unsupervised deep clustering methods with the aid of contrastive representation learning have been proposed for computer vision \cite{caron2018deep, zhong2020deep, li2021contrastive} and natural language processing \cite{zhang2021supporting}. To decompose the intrinsic multimodal distribution of demonstrations, we also adopt the intuition of jointly learning representations and clustering in an end-to-end manner. While prior works either focus on visual representations in the form of a single fixed-length vector \cite{zhong2020deep, li2021contrastive} or rely on data augmentations with a heavy computational burden \cite{zhang2021supporting}, our Sequence-based Contrastive Clustering is unique in handling trajectory data with various sequential lengths, by utilizing simple and efficient down-sampling to construct positive pairs of contrastive learning.
}

\vspace{-2pt}
\section{Out-of-dynamics Imitation Learning}
\vspace{-2pt}

In \OODIL, we aim to learn a policy for the agent of interest (the imitator) from demonstrations collected from multiple demonstrators with different dynamics from the imitator. Formally, we model both the demonstrators and the imitator as a standard Markov decision process (MDP) $\mathcal{M}= \langle \mathcal{S}, \mathcal{A}, p, \mathcal{R}, \rho_0, \gamma \rangle$. Here $\mathcal{S}$ is the state space, $\mathcal{A}$ is the action space, $p: \mathcal{S} \times \mathcal{A} \times \mathcal{S}\rightarrow [0,1] $ is the transition probability and $\mathcal{R}: \mathcal{S} \times \mathcal{S} \rightarrow \mathbb{R}$ is the reward function, which is solely defined on states and shared between the demonstrators and the imitator. Such design aims to satisfy the basic requirement of imitation learning that the demonstrators and the imitator should finish the same task~\cite{liu2019state,cao2021learning}. $\gamma$ is the shared discount factor. The action spaces $\mathcal{A}$ and the transition probability $p$ of different demonstrators and the imitator are different. In particular, we use $\mathcal{M}^i= \langle \mathcal{S}, \mathcal{A}^i, p^i, \mathcal{R}, \rho_0, \gamma \rangle$ to indicate the MDP of the imitator. A policy for the imitator $\pi^i: \mathcal{S} \times \mathcal{A}^i \rightarrow [0,1]$ defines a probability distribution over the action space in a given state. The policy is evaluated by the expected return, which is defined by $\eta_{\pi^i}= \mathbb{E}_{s_0 \sim \rho_0,\pi^i}\left[\sum_{t=0}^{\infty}\gamma^{t} \mathcal{R}(s_{t}, s_{t+1})\right]$, where $t$ indicates the time step.

We formalize the policy learning as an imitation learning problem where the reward function $\mathcal{R}$ is unknown. We aim to learn the policy $\pi^i$ from a set of demonstrations collected by different demonstrators $\Xi=\{\xi_1, \xi_2, \dots\}$ where each trajectory is a sequence of states $\xi=\{s^d_0, s^d_1,\dots,s^d_H\}$. Here we use state trajectories since different action spaces make it impossible to imitate actions. Since we put no assumption on the dynamics of the demonstrators, there is no guarantee that an optimal policy could be learned, e.g., in the extreme case, all the demonstrations could be non-transferable and a random policy will be learned. Our goal is to learn a policy with as high return as possible.

Since the dynamics of the demonstrators and the imitator are different, the transitions $(s^d_t, s^d_{t+1})$ in the demonstrations may not be realizable by the imitator, i.e., no action $a_t^i$ in the imitator's action space $\mathcal{A}^i$ could make $p^i(s^d_t,a_t^i,s^d_{t+1})>0$. Such demonstrations provide no useful information for the imitator and are non-transferable. However, given such a set of demonstrations from a mixture of demonstrators, it is non-trivial to directly test whether a trajectory is achievable by reproducing the trajectory in the target environment since there is no action in demonstrations. Furthermore, different demonstrators may take different policies due to different dynamics and may take different actions even at the same state\footnote{This work mainly focuses on the case where the dynamics are nearly-deterministic like robotic applications so unimodal policies can rarely realize multimodal demonstrations.}. This multimodal distribution of the demonstrations makes prior works on transferability measurement ineffective. In this paper, we aim to remove the interference from the multimodal distribution of the demonstrations and learn transferability to mitigate the negative impact of non-transferable demonstrations.

\vspace{-2pt}
\subsection{Sequence-based Contrastive Clustering}
\vspace{-2pt}

Since directly imitating from a mixture of demonstrations will make a unimodal policy suffer, we need to fully capture and decompose the intrinsic multimodal distribution of data collected from different demonstrators. Thus, we cluster demonstrations before learning their transferability to make sure they belong to a unimodal distribution. 
Motivated by contrastive learning~\cite{chen2020simple}, we learn a sequence feature extractor and a distance metric in the sequence feature space by contrastive learning over carefully-designed positive and negative pairs. We construct positive pairs by randomly subsampling fixed-length sub-trajectories from the same demonstration and treat sub-trajectories of different demonstrations as negative pairs. Such a design satisfies our requirement since the positive pairs are from the same trajectory, and thus are from the same mode. At each batch, we first sample $N$ trajectories from $\Xi$ and for each trajectory $\xi_i$, we take two sub-trajectories $\xi^{\text{sub}}_{2i-1}$ and $\xi^{\text{sub}}_{2i}$. We then derive the contrastive learning loss for this batch of $2N$ sub-trajectories as follows:
\begin{equation}
    \mathcal{L}_{\text{contrast}}= -\frac{1}{N}\sum_{i=1}^{N}\log \frac{\exp(\langle F(\xi^{\text{sub}}_{2i-1}), F(\xi^{\text{sub}}_{2i})\rangle)}{\sum_{i'\ne 2i-1, i'\ne 2i} \exp(\langle F(\xi^{\text{sub}}_{2i-1}), F(\xi^{\text{sub}}_{i'})\rangle)+\exp(\langle F(\xi^{\text{sub}}_{2i-1}), F(\xi^{\text{sub}}_{2i})\rangle)},
\end{equation}
where $F$ is the feature extractor modeled as a recurrent neural network and $\langle \cdot ,\ \cdot \rangle$ indicates the cosine distance. The loss maximizes the cosine similarity of features between positive pairs.

We embed contrastive learning into clustering to simultaneously ensure that the positive pairs are close and the clustering structure is learned. We first initialize a matrix $\mathbf{C}$ of size $\mathrm{dim}(F)\times K$, where $\mathrm{dim}(F)$ is the dimension of the output of $F$ and each column $\mathbf{c}_k$ of $\mathbf{C}$ represents the center of cluster $k$, with $K$ clusters in total. For each input sub-trajectory $\xi^{\text{sub}}_n$, we assign a one-hot cluster label $\mathbf{y}_n\in \{0,1\}^K$ as follows:
\begin{equation}\label{eqn:assignment}
    y_{n,j} = \begin{cases}
    1 , \quad j = \arg\min_{k={1,\cdots,K}} \lVert F(\xi^{\text{sub}}_n) - \mathbf{c}_k \rVert_2 \\
    0 , \quad \text{otherwise} \\
    \end{cases}
\end{equation}
where we assign the cluster label with the nearest cluster center to a sub-trajectory. Then we introduce the clustering objective with the contrastive learning loss $\mathcal{L}_{\text{contrast}}$ embedded into it:
\begin{equation}\label{eqn:cluster_update}
    \mathcal{L}_{\text{cluster}} = \mathcal{L}_{\text{contrast}} + \frac{\lambda}{2} \lVert F(\xi^{\text{sub}}_n) - \mathbf{C}\mathbf{y}_n \rVert_2^2.
\end{equation}
Every time we update the feature extractor $F$, we reassign the cluster label of each sub-trajectory by Eqn.~\eqref{eqn:assignment} and update the cluster center by the following process:
\begin{equation}\label{eqn:update}
    \mathbf{c}_k \leftarrow (1-\beta) \mathbf{c}_k + \beta \sum_{n=1}^{2N} \mathbb{I}[y_{n,k} = 1] F(\xi^{\text{sub}}_n),
\end{equation}
where $\mathbb{I}$ is an indicator taking $1$ only when the condition is satisfied and $0$ otherwise, and $\beta=\frac{1}{\sum_{n=1}^{2N} \mathbb{I}[y_{n,k} = 1]} $ automatically controls the learning rate. We simultaneously optimize the contrastive learning objective and the clustering objective to encourage them to benefit from each other, where the cluster structure is learned to keep trajectories from the same mode to be in the same cluster. Our contrastive clustering algorithm derives $K$ clusters of trajectories $\Xi_k|_{k=1}^{K}$, where $\Xi_k$ contains trajectories with cluster label $k$. We show the algorithm procedure in Appendix.

\begin{figure}
    \centering
    \includegraphics[width=.9\textwidth]{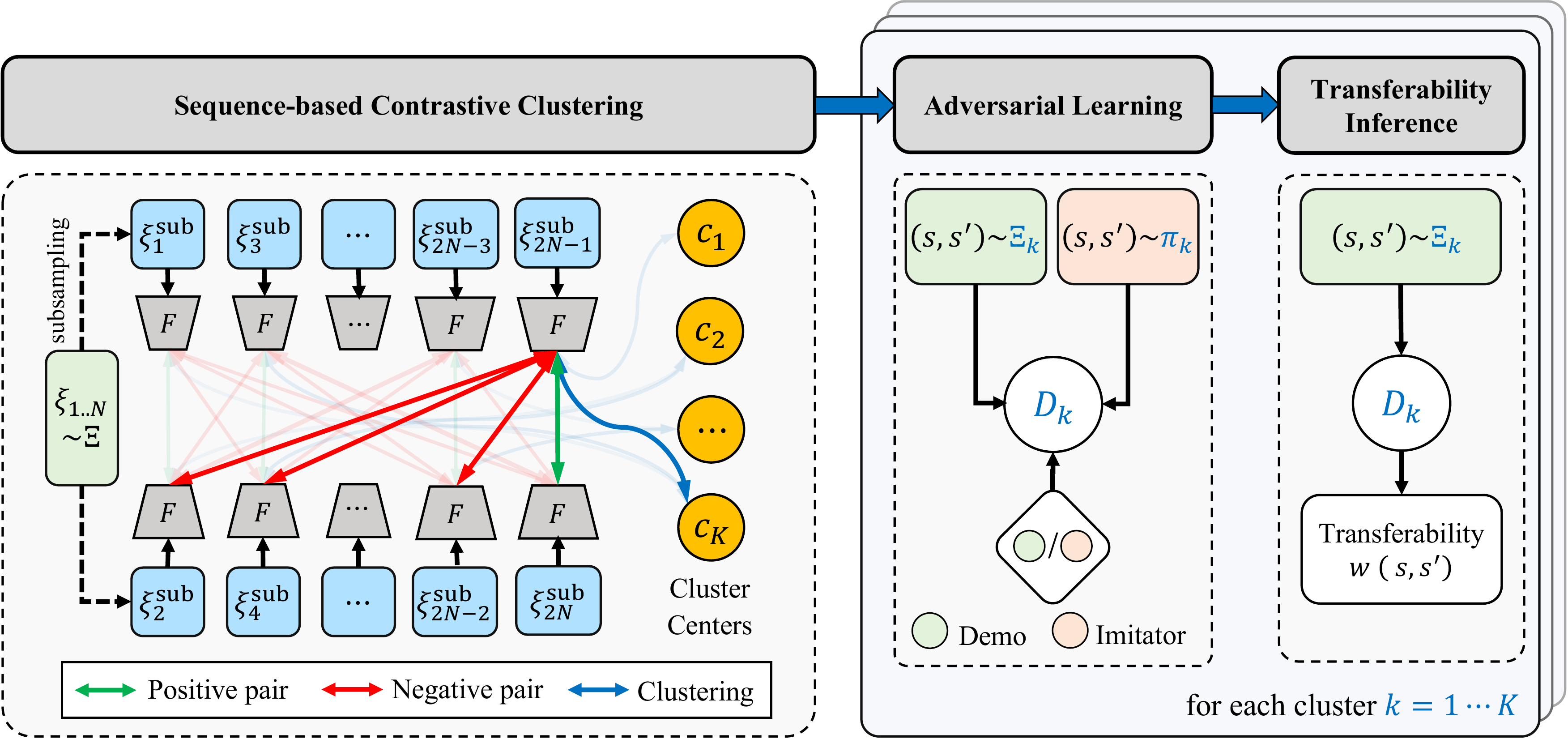}
    \caption{The outline of our whole algorithm can be divided into two phases. The first phase is sequence-based contrastive clustering where we simultaneously conduct contrastive learning and clustering. We create positive pairs by subsampling different sub-trajectories from the same trajectory and use sub-trajectories from different trajectories as negative pairs. The second phase is learning transferability where we conduct an adversarial-learning based algorithm in each cluster.}
    \label{fig:outline}
    \vspace{-10pt}
\end{figure}

\vspace{-2pt}
\subsection{Adversarial Transferability Measurement}
\vspace{-2pt}

Given that the sequence-based contrastive clustering algorithm removes the mutual interference of demonstrations from different modes, we then learn the transferability within each cluster.
We design a new transferability measurement by an adversarial-learning algorithm based on generative adversarial imitation learning (GAIL)~\cite{ho2016generative}. Our key insight is that the GAIL discriminator output indicates the likelihood for a state transition to either come from the demonstration distribution or the policy distribution. 
We train a GAIL policy $\pi_k$ and a GAIL discriminator $D_k$ for each cluster of trajectories. The loss is defined as follows:
\begin{equation}
  \mathcal{L}_{\text{tran}} = - \sum_{k=1}^{K}\left(\mathbb{E}_{(s^d_t,s^d_{t+1})\sim \Xi_k}\log (1-D_k(s^d_t,s^d_{t+1})) + \mathbb{E}_{(s^{\pi_k}_t,s^{\pi_k}_{t+1})\sim \pi_k}\log (D_k(s^{\pi_k}_t,s^{\pi_k}_{t+1}))\right).  
\end{equation}
Here $\pi_k$ indicates the current policy in training and $D_k$ indicates the discriminator. We use label $0$ for the state transitions in the demonstrations and $1$ for the state transitions collected from the policy.

After convergence, the discriminator for each mode outputs a value in $[0,1]$ reflecting how likely an input transition is drawn from the state transition distribution derived by the imitator's policy. Thus for a state transition in the demonstrations, if it has a discriminator output close to $1$, it is more possible to be drawn from the imitator's policy and achievable by the imitator. Thus, we quantify the transferability for a state transition as follows:
\begin{equation}\label{eqn:feasibility}
    w(s^d_t, s^d_{t+1}) = \sum_{k=1}^{K} \mathbb{I}\left[(s^d_t, s^d_{t+1})\in \Xi_k\right]D_k(s^d_t, s^d_{t+1}).
\end{equation}
Note that the discriminator is strengthened by the adversarial training paradigm and thus has a strong ability to discriminate whether a transition is from the demonstrations or the imitator's policy. The optimization of the discriminator is demonstrated to be efficient and effective~\cite{ho2016generative}, which requires no time-consuming step-by-step optimization in f-MDP~\cite{cao2021corl}.

\noindent \textbf{Transferablity-sampling Imitation Learning.}
We finally embed the transferability quantified by Eqn.~\eqref{eqn:feasibility} into imitation learning. We normalize the transferability of all the state transitions into a sampling distribution, where state transitions with larger transferability will be sampled more often: 
\begin{equation}
    p_w(s^d_t, s^d_{t+1})=\frac{w(s^d_t, s^d_{t+1})}{\sum_{(s^d_{t'}, s^d_{{t'}+1})\in \Xi}w(s^d_{t'}, s^d_{{t'}+1})}.
\end{equation}
\textcolor{revisioncolor} {Using the fixed sampling distribution $p_w$, we can embed our transferability into any imitation-learning-from-observations algorithm~\cite{torabi2018behavioral, ho2016generative, fu2017learning}.} We use GAIL~\cite{ho2016generative} here to train the final policy:
\begin{equation}
    \mathcal{L}_{\text{GAIL}} = -\mathbb{E}_{(s^d_t,s^d_{t+1})\sim p_w}\log (1-D(s^d_t,s^d_{t+1})) - \mathbb{E}_{(s^\pi_t,s^\pi_{t+1})\sim \pi}\log (D(s^\pi_t,s^\pi_{t+1})).
\end{equation}

\vspace{-10pt}
\section{Experiments}
\vspace{-5pt}

We experiment with three MuJoCo environments, a simulated Driving environment, and a simulated Franka Panda Arm environment. We compare our approach with a standard imitation learning algorithm: GAIL~\cite{ho2016generative}, imitation learning with a measure of feasibility: ID~\cite{cao2021learning} and f-MDP~\cite{cao2021corl}. The original ID learns the inverse dynamics model from random trajectories far from demonstrations and thus makes the learned inverse dynamics not work well on demonstrations. We create an advanced ID baseline by learning a GAIL policy to generate trajectories as the data for inverse dynamic learning. Such trajectories are closer to demonstrations and make the inverse dynamics work better on demonstrations. We call the original ID as ID-Random and the advanced ID as ID-GAIL.

We further conduct analyses including an ablation study to verify the efficacy of each component and include a visualization of the learned transferability, the results for different compositions of demonstrations, and the performance gain when we are given a larger budget to collect demonstrations in the Appendix. Code is available at \url{https://github.com/EvieQ01/OODIL}. 

\vspace{-5pt}
\subsection{MuJoCo}\label{sec:mujoco}
\vspace{-5pt}

\begin{figure}[ht]
    \centering
    \includegraphics[width=.9\textwidth]{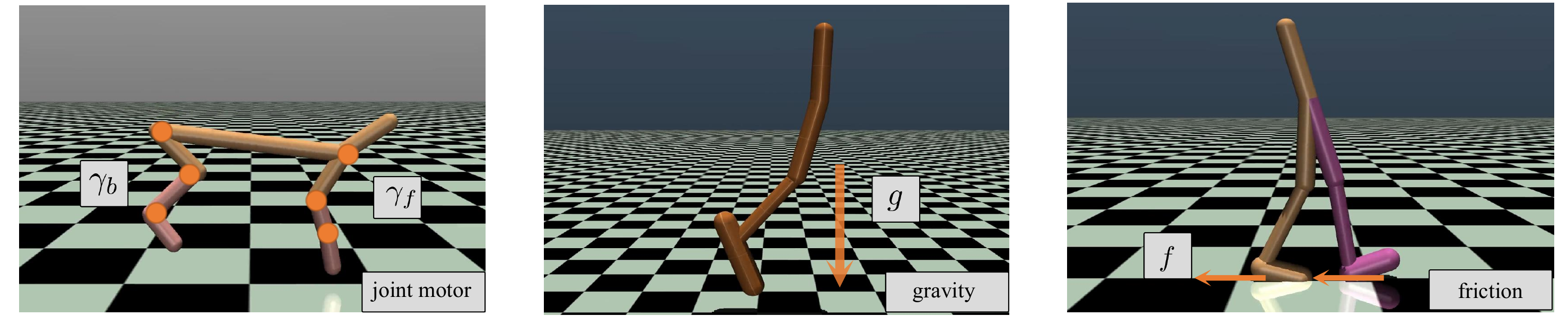}
    \caption{The illustration of the varying dynamics for HalfCheetah, Hopper, and Walker2d.}
    \label{fig:mujoco_illus}
\end{figure}
\vspace{-5pt}

\noindent \textbf{Environments.} We illustrate the environments in Figure~\ref{fig:mujoco_illus}. The \textbf{HalfCheetah} is an agent with two legs and a body. We create different dynamics by discounting the force of the front leg and the back leg with two factors $\gamma_f$ and $\gamma_b$ respectively. We create a mixture of demonstrators by setting $(\gamma_f, \gamma_b)$ as (i) $(1, 0.9)$, (ii) $(0.9, 1)$, (iii) $(1, 0.05)$, (iv) $(0.05, 1)$. The imitator has the original force: $(\gamma_f, \gamma_b)$ as $(1, 1)$. The \textbf{Hopper} is an agent with one leg consisting of $3$ joints. We create different dynamics by varying the gravitational constant. We create a mixture of demonstrators by setting the gravitational constant as (i) $15.0$, (ii) $9.8$, (iii) $2.0$, (iv) $1.0$. For the imitator, we use a gravitational constant of $12.0$. The \textbf{Walker2d} is an agent with two legs where each leg consists of $3$ joints. We create different dynamics by using different frictions for the feet, i.e., the link that touches the ground. We create a mixture of demonstrators by setting the friction as (i) $24.8$, (ii) $9.9$, (iii) $3.9$, (iv) $1.1$. For the imitator, we use a friction of $19.9$. For all three MuJoCo environments, we follow prior works~\cite{cao2021corl,cao2021learning} to collect 
fewer demonstrations from more transferable environments. We collect $100, 100, 250, 250$ demonstrations respectively for four dynamics in HalfCheetah, $10, 10, 250, 250$ for Hopper and $25,50,50,50$ for Walker2d with $1000$ interaction steps per demonstration.  \textcolor{revisioncolor}{We train expert agents with Trust Region Policy Optimization (TRPO) \cite{schulman2015trust} to generate demonstrations.}
\begin{figure}[t]
\centering
\vspace{-5pt}
    \includegraphics[width=.8\textwidth]{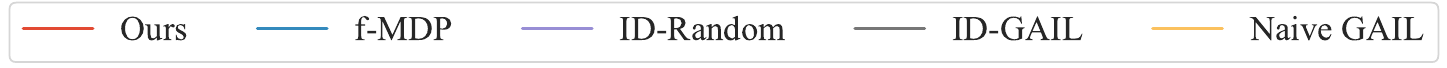}
\end{figure}

\begin{figure}[t]
\centering
\vspace{-5pt}
    \subfigure[HalfCheetah]{\includegraphics[width=0.33\textwidth]{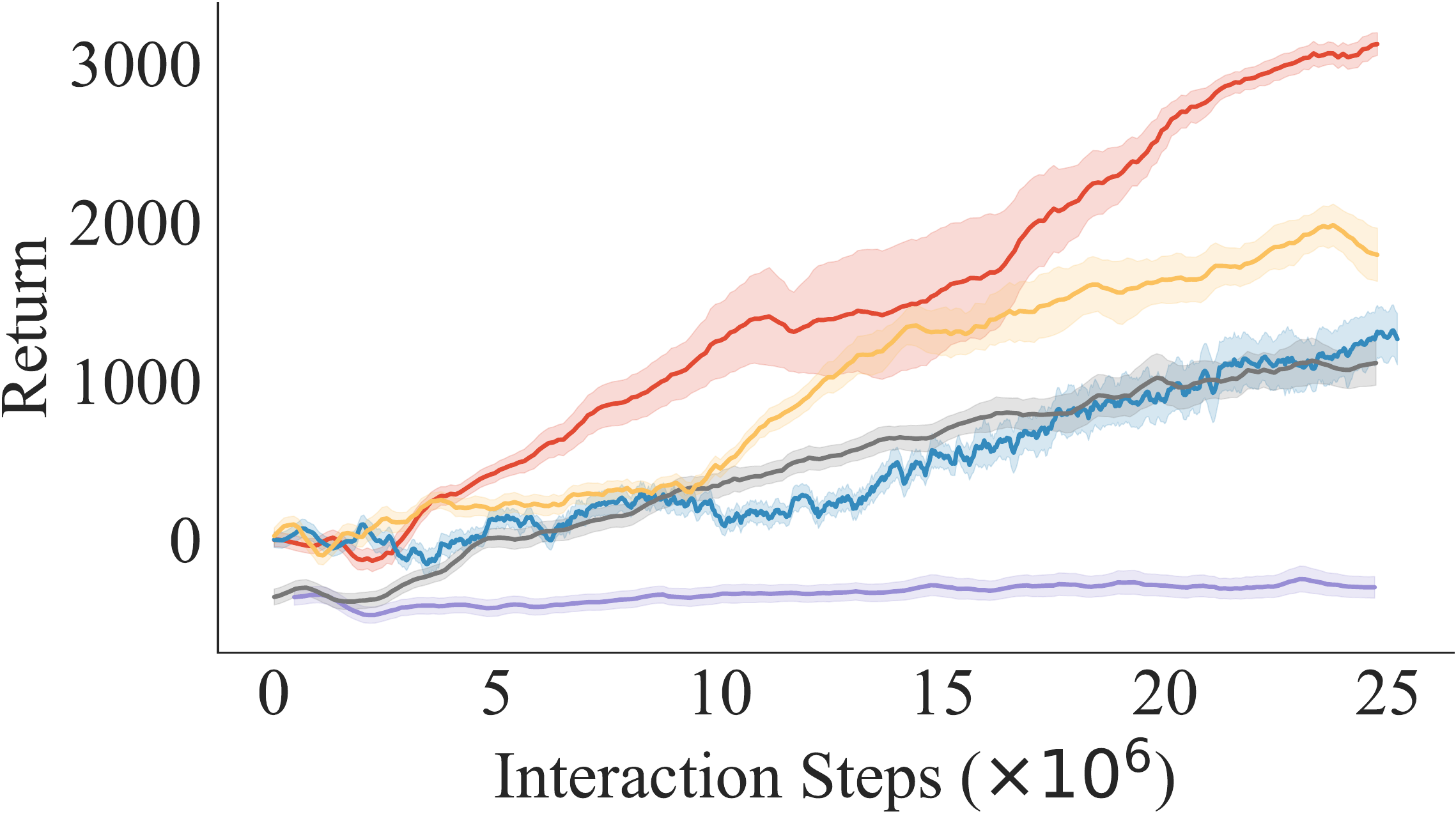}\label{fig:cheetah}}
    \subfigure[Hopper]{\includegraphics[width=0.33\textwidth]{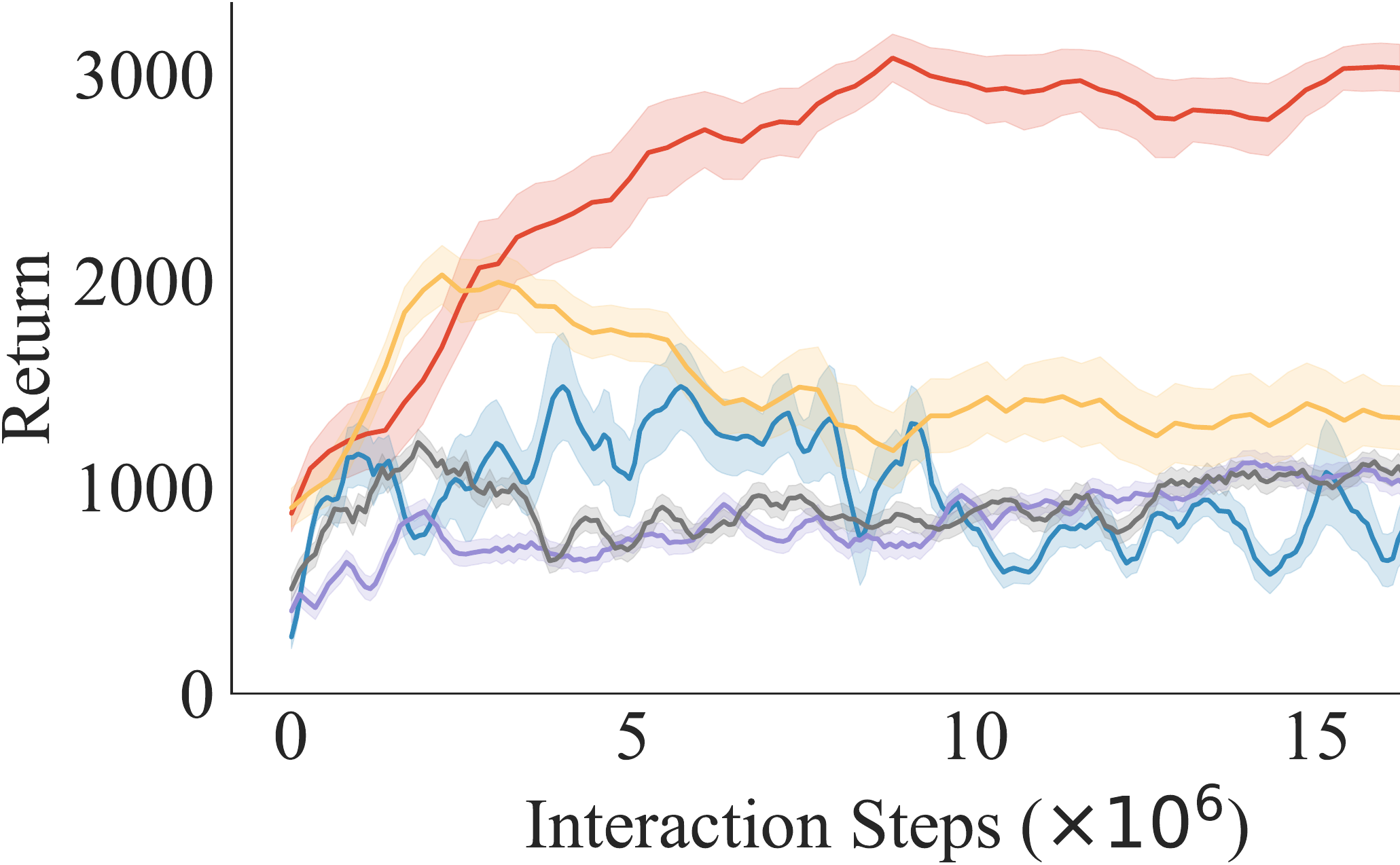}\label{fig:hopper}}
    \subfigure[Walker2d]{\includegraphics[width=0.32\textwidth]{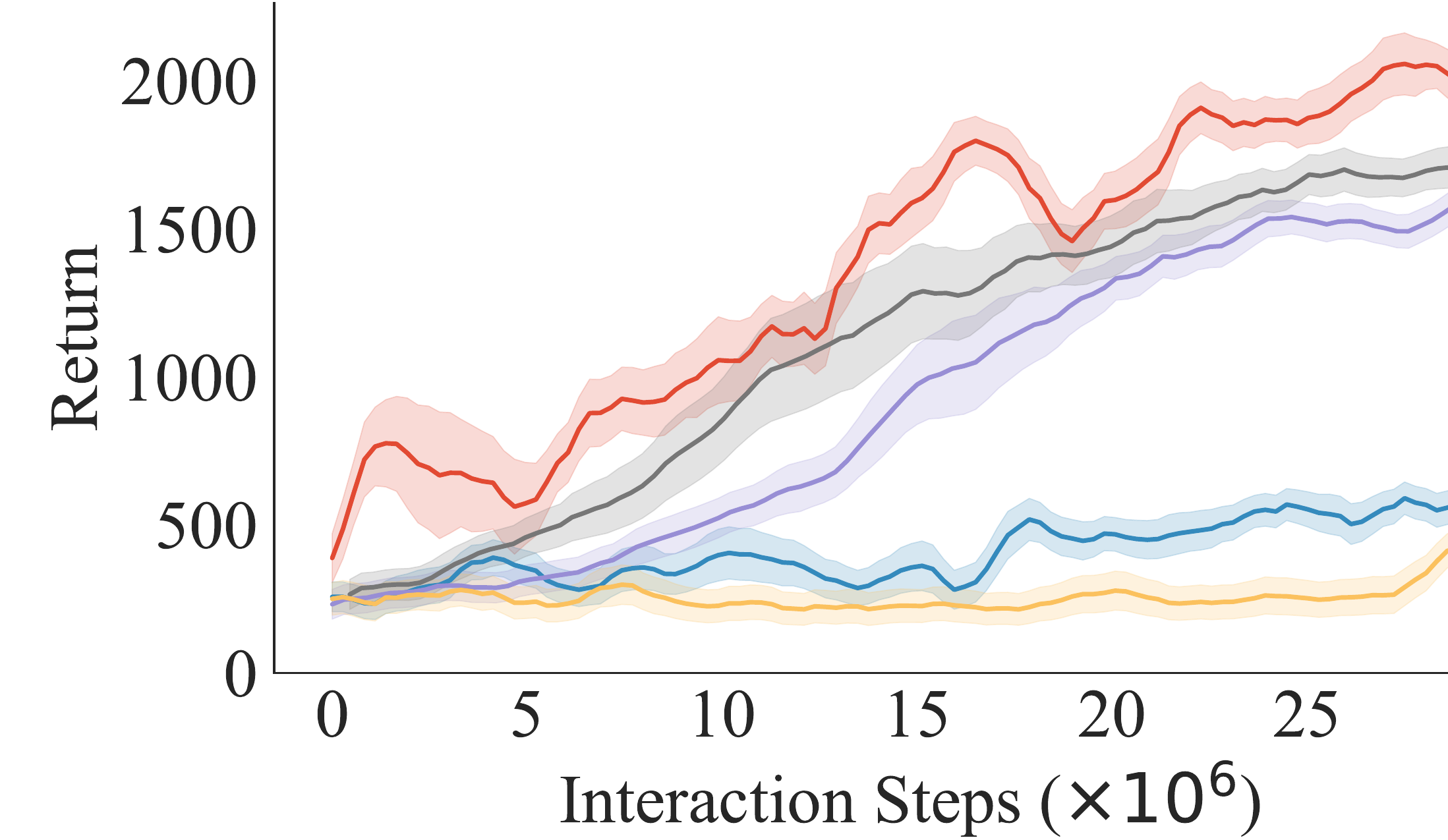}\label{fig:walker2d}}
    \vspace{-5pt}
    \caption{The imitation learning performance of different methods in the three MuJoCo environments.}\label{fig:mujoco_result}
    \vspace{-10pt}
\end{figure}

\noindent \textbf{Results.} We show the expected return w.r.t. the number of interaction steps for the three environments in Figure~\ref{fig:mujoco_result}. In all three environments, we observe that the proposed method achieves the highest return. The baselines even show lower performance than naive GAIL under such multimodal distribution of source demonstrations because f-MDP learns a unimodal policy from the multimodal trajectory distribution, which cannot realize the demonstration trajectories, while ID fails to learn an accurate inverse dynamics model from random trajectories. The results demonstrate the importance of clustering trajectories of the same mode. 
By learning the transferability in each cluster, we can accurately filter out non-transferable demonstrations and learn from transferable demonstrations.

\vspace{-5pt}
\subsection{Driving}
\vspace{-2pt}

\begin{figure}[h]
\vspace{-5pt}
    \centering
    \subfigure[Illustration]{\includegraphics[width=0.42\textwidth]{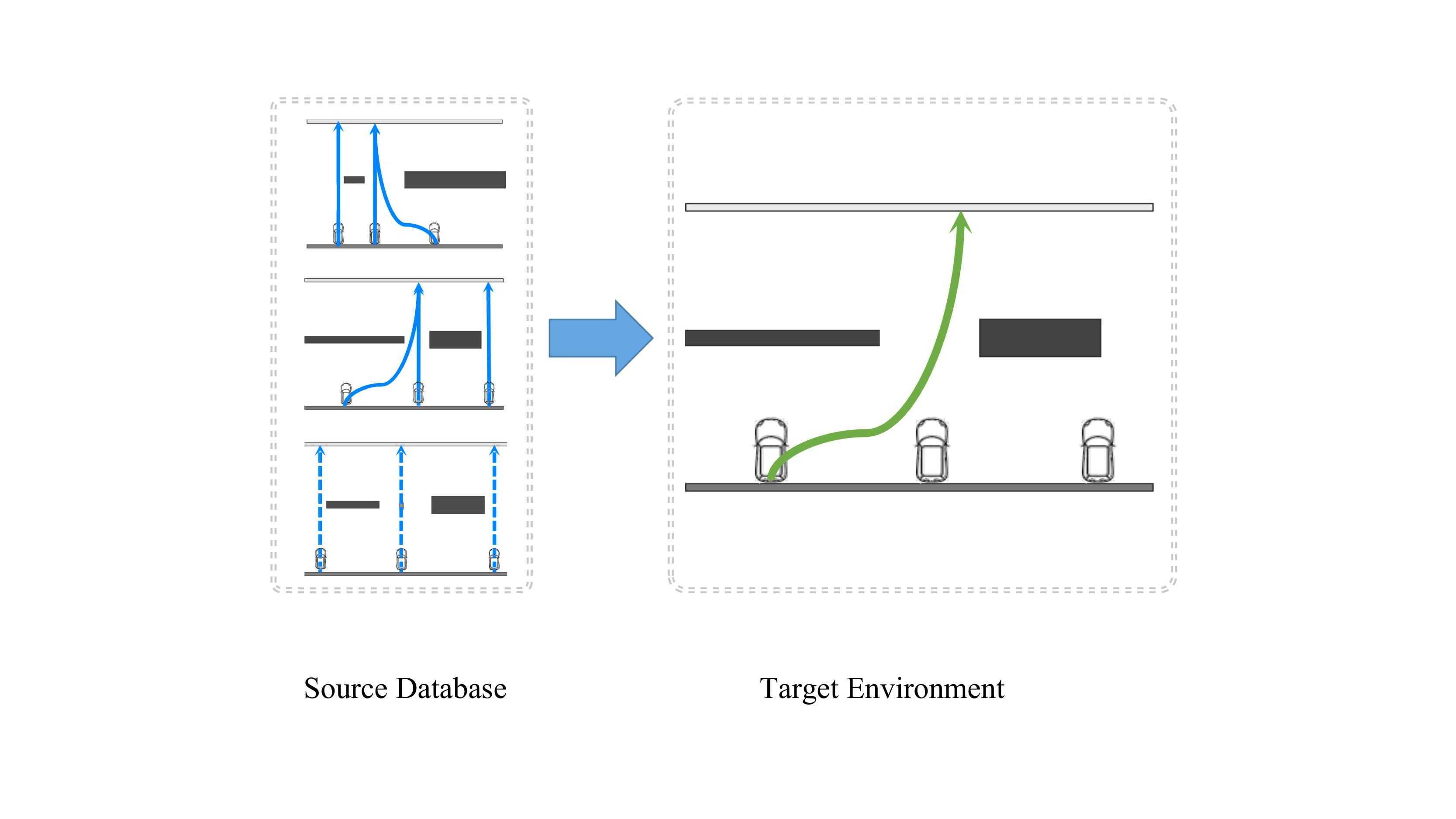}\label{fig:driving_illustration}}
    \subfigure[Result]{\includegraphics[width=0.41\textwidth]{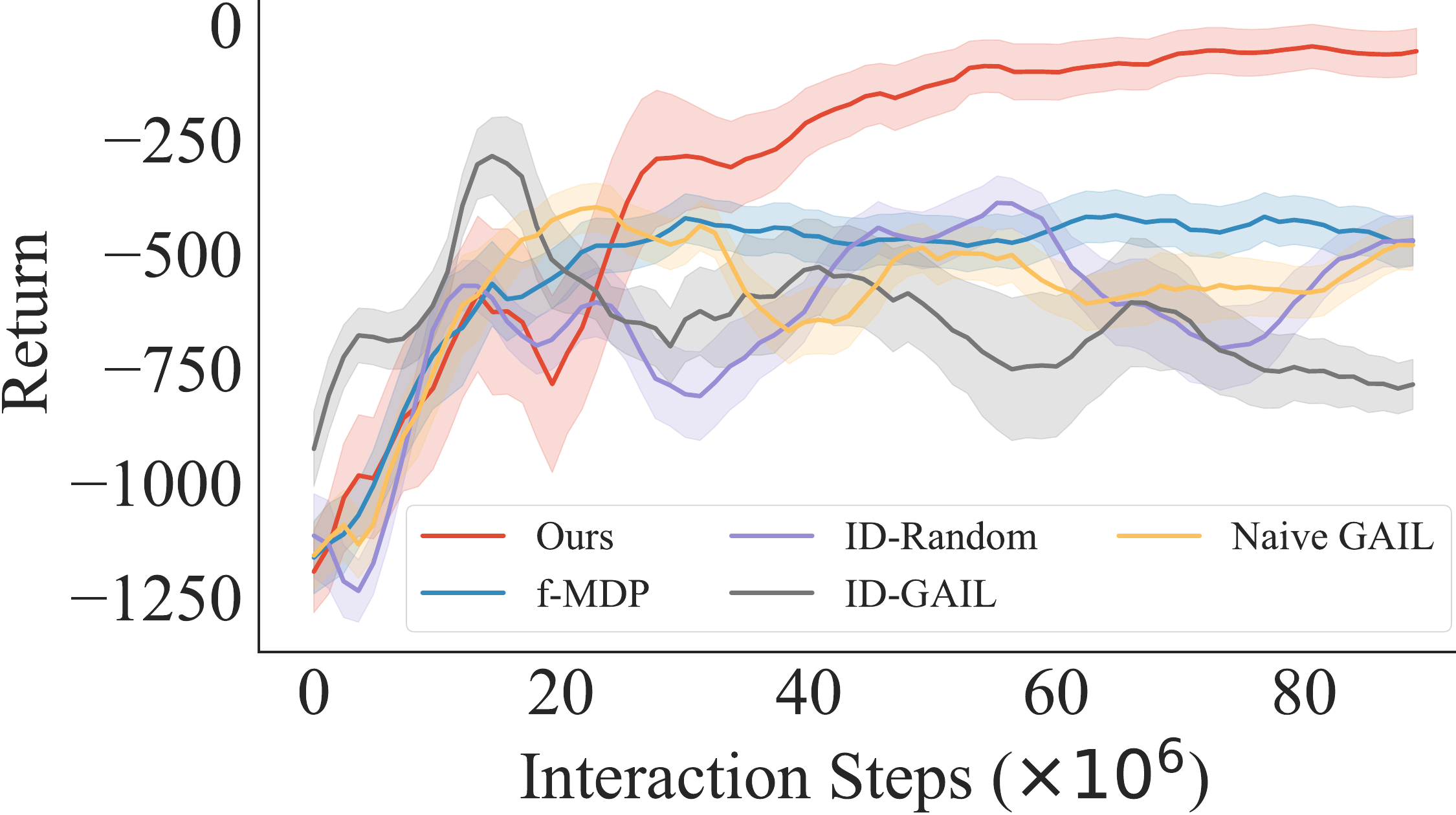}\label{fig:driving_result}}
    \vspace{-5pt}
    \caption{(a) The illustration of the driving environment, showing expert trajectories in multiple source dynamics and the target environment. (b) The results in the driving environment.}
    \label{fig:driving}
    \vspace{-2pt}
\end{figure}

\noindent \textbf{Environment.} In the driving environment, we can easily decide and interpret whether the target car can reproduce the route in the demonstrations. As shown in Figure~\ref{fig:driving_illustration}, we create a task where a car drives starting from anywhere at the bottom side and ends at the top side. Two obstacles are set with the center at $\frac{1}{4}$ width and $\frac{3}{4}$ width respectively. We create different dynamics by setting obstacles with different widths and setting different speeds for the car, which simulates a realistic scenario where different car models are driving at different places. The reward function is defined as $-1$  for each interaction step, $+1000$ for reaching the goal, and $-1000$ for hitting the obstacle. In this environment, the different lengths and paths of the demonstration trajectories introduce a clear multimodal distribution, which can demonstrate the importance of clustering trajectories into an accurate mode. We create three demonstrators by setting the obstacles width as $[0.1, 0.5]$, $[0.5, 0.25]$ and $[0.25, 0.25]$ and setting the speed as $1.0$, $1.0$ and $5.0$ respectively. For the target environment, we set the obstacle width as $[0.4, 0.25]$ and the speed as $1.0$. We collect $5\times10^5$, $2.5\times10^6$, and $2.5\times10^6$ interaction steps of demonstrations from each source dynamics, by handmade rules.

\noindent \textbf{Results.} As is shown in Figure~\ref{fig:driving_result}, we observe that the proposed method outperforms all other baselines. Though ID performs well at the first few interaction steps but the performance drops much then, which can be explained by that the feasibility learned by ID could find good demonstrations in the first few batches of data but makes errors then. That means ID can only learn partially correct feasibility. Instead, our method learns accurate transferability to filter out non-transferable demonstrations and converges stably to a high return. f-MDP, as the state-of-the-art method for filtering non-transferable demonstrations, could not work well on this multimodal distribution of demonstrations, further indicating the importance of clustering trajectories into the correct mode.

\vspace{-5pt}
\subsection{Simulated Franka Panda}
\vspace{-2pt}

\begin{figure}[t]
    \centering
    \subfigure[Illustration]{\includegraphics[width=0.5\textwidth]{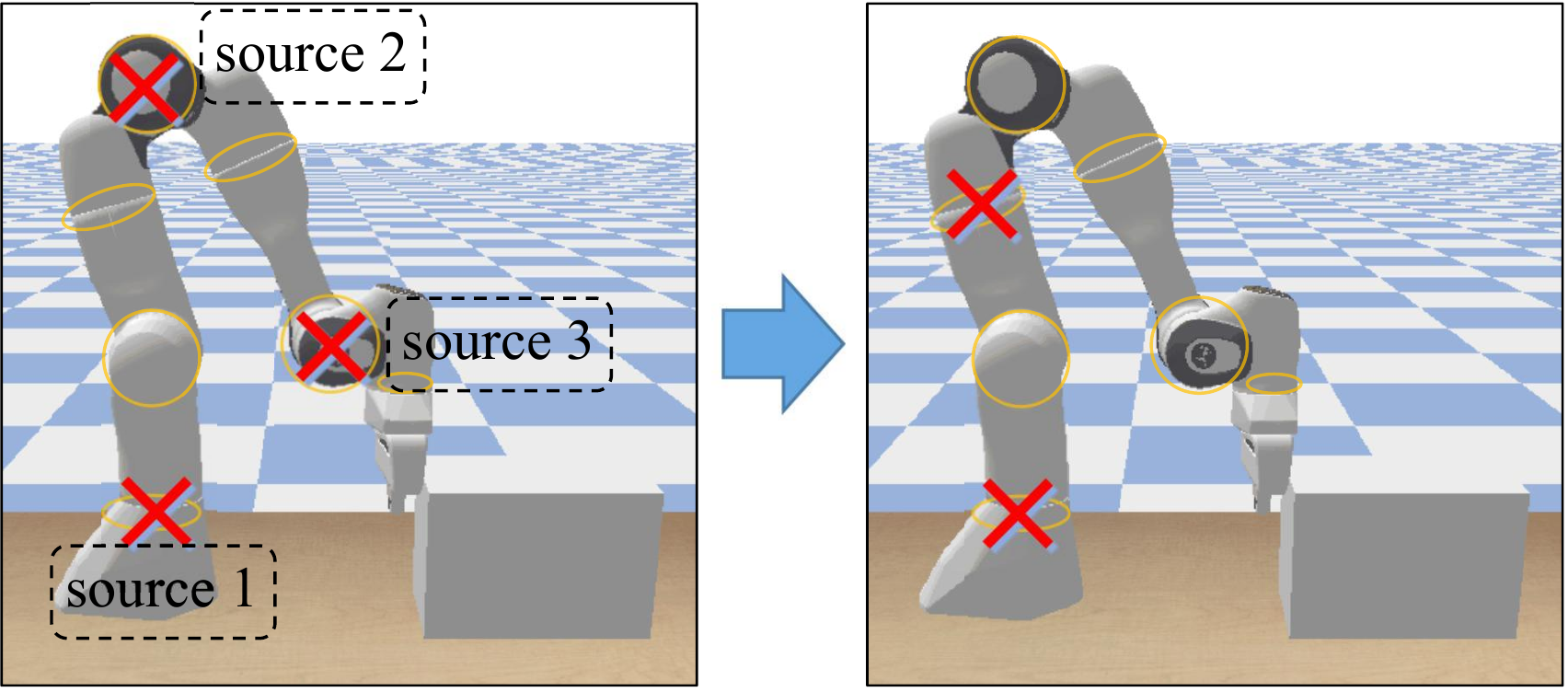}\label{fig:panda_illustration}}
    \subfigure[Result]{\includegraphics[width=0.4\textwidth]{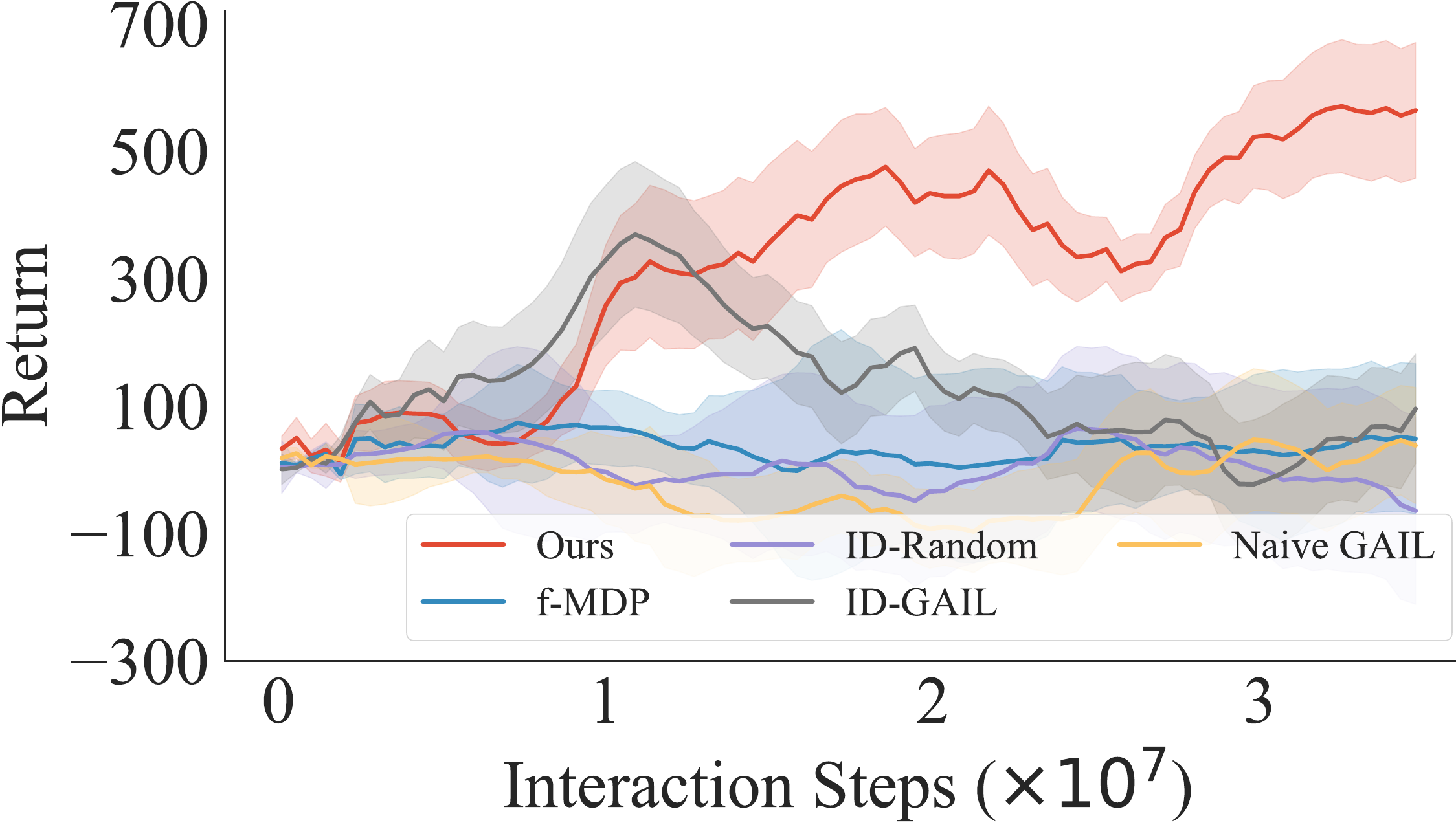}\label{fig:panda_result}}
    \vspace{-5pt}
    \caption{(a) The illustration of Simulated Franka Panda environment, where we create different demonstrators by disabling different single joints and disabling two joints for the imitator. (b) The return of different methods in the Simulated Franka Panda environment.}
    \label{fig:panda}
    \vspace{-10pt}
\end{figure}

\noindent \textbf{Environment.} The environment simulates the Franka Panda Robot Arm\footnote{https://www.franka.de/} with $7$ degrees of freedom (DoF), which is implemented in the PyBullet~\cite{coumans2016pybullet}. \textcolor{revisioncolor}{We create a task of pushing a box from one side of the desk to the other. With the box set at a base position and the robot arm set at a random position with Gaussian distribution, we set the target at the right of the desk}. We create different dynamics by disabling different joints of the Robot arm. As shown in Figure~\ref{fig:panda_illustration}, we create three demonstrators by disabling the No. $1$, $4$, and $6$ joint respectively while disabling the No. $1$ and $3$ joints for the target imitator. The environment aims to address a real problem to leverage historical data on any robot to learn a policy for a new robot. Similar to the MuJoCo environment, we import more demonstrations from the more dissimilar demonstrator, where the number of interaction steps is $1\times 10^6$, $1\times 10^5$, and $1\times 10^6$ for the environment disabling No. $1$, $4$, and $6$ respectively. The reward function is defined as the current distance to the starting point, and an extra $+1000$ for the box reaching the other side of the table and $-1000$ for the box dropping to the ground or the robot going past the box to the other side of the desk. \textcolor{revisioncolor}{We make demonstrations by handmade rules.}

\noindent \textbf{Results.}
The expected return w.r.t. the number of interaction steps is shown in Figure~\ref{fig:panda_result}. Our proposed method outperforms all the baselines by a large margin. Directly applying GAIL introduces a low performance, which indicates that the demonstrations from other robots cannot be directly utilized to learn a new robot. The experiments show that the proposed method can serve as a data cleaning step to clean the dataset from other robots for a real-robot transfer learning problem.

\vspace{-5pt}
\subsection{Analysis}
\vspace{-15pt}

\noindent
\begin{wrapfigure}{r}{0.35\textwidth}
    \vspace{5pt}
    \centering
     \includegraphics[width=0.35\textwidth]{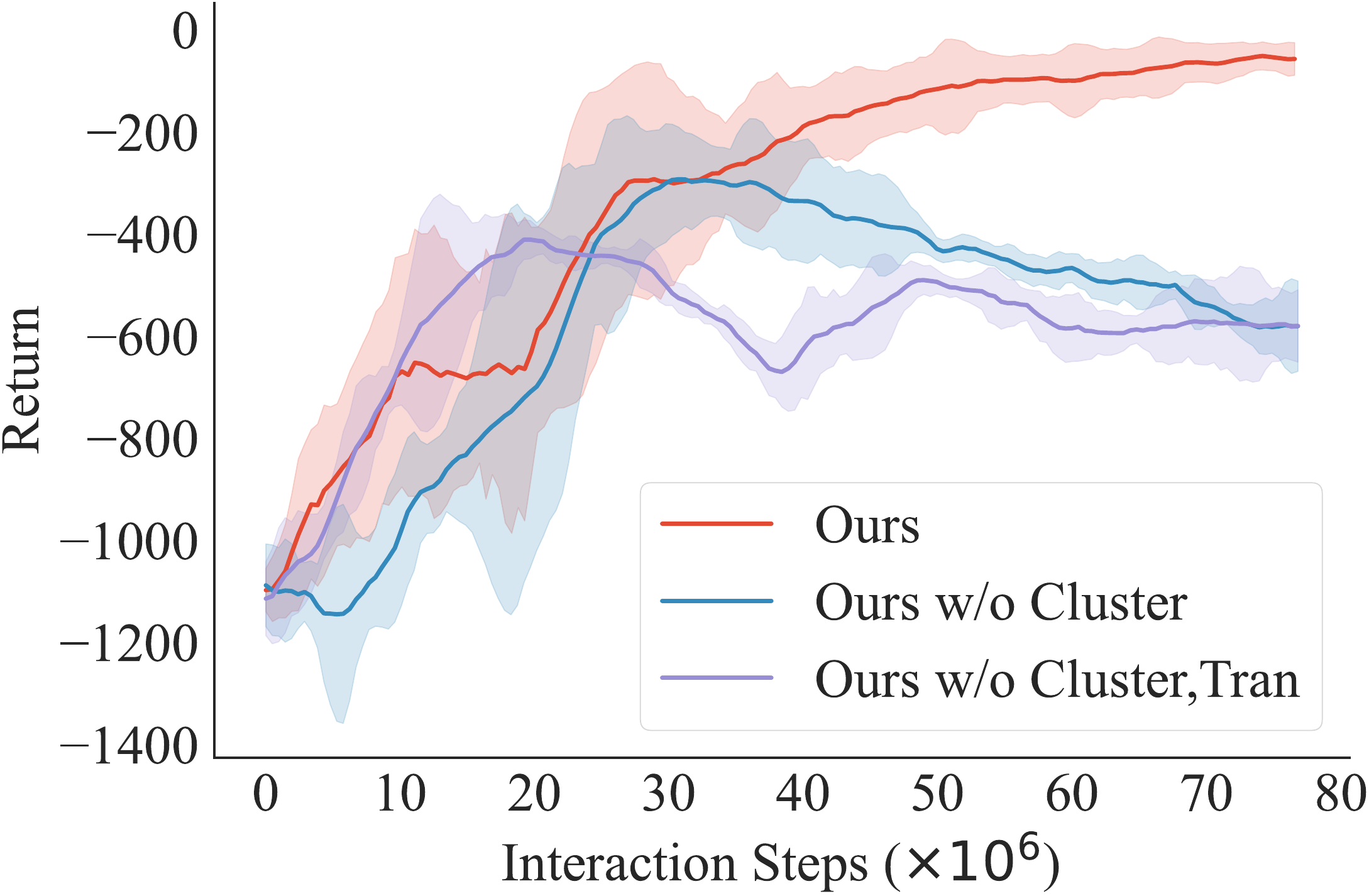}
    \caption{The ablation study on two variants of our method.}
    \label{fig:ablation}
    \vspace{-20pt}
\end{wrapfigure}

\textbf{Ablation Study.} 
To verify that both our contrastive clustering algorithm and the adversarial-learning based algorithm contribute to the final performance, 
we compare the performance of our method with its variants by removing the clustering step and learning the transferability directly from the whole set of demonstrations (Ours w/o Cluster), and removing both clustering and the transferability (Ours w/o Cluster, Tran), which directly imitates the whole set of demonstrations. \textcolor{revisioncolor}{We conduct these experiments in the Driving environment.}

The results are shown in Figure~\ref{fig:ablation}. We observe that Ours outperforms Ours w/o Cluster, which demonstrates that clustering trajectories within the same mode is important and our contrastive clustering algorithm achieves this goal. Ours w/o Cluster outperforms Ours w/o Cluster, Tran, which demonstrates that transferability is important to filter out non-transferable demonstrations and learn from more transferable ones.

\vspace{-5pt}
\section{Conclusion}
\vspace{-5pt}

We propose a new approach to address out-of-dynamics imitation learning (OOD-IL). Noticing that the demonstrations exhibit a multimodal distribution, we propose a sequence-based contrastive clustering algorithm to make trajectories from the same mode fall into the same cluster. We then propose an adversarial-learning based algorithm to learn the transferability of each cluster with the discriminator output. Experimental results on three MuJoCo environments, a driving environment, and a simulated robot environment show that the proposed method can learn a transferability measure to accurately filter out non-transferable demonstrations and learn from more transferable 
ones.

\textbf{Limitations and future work.} \textcolor{revisioncolor}{
While our work has substantially advanced OOD-IL, we believe that this problem merits further study. 
One primary limitation of our current method is that our method may not afford the computational cost of transferability learning when the number of clusters increases to hundreds or thousands. Pre-training or meta-training \cite{finn2017model} before transferability learning may help boost learning efficiency, which is left for future work.
A limitation of deploying our method into a real system is that it may not respect safety constraints due to exploration during both transferability learning and imitation learning. Although this challenge is common for real systems, out-of-dynamics learning may exacerbate it and safe policy learning techniques can be adopted.
Another promising future direction is to exploit a wider variety of demonstrations, thus enabling large-scale imitation learning. Although our work relaxes the stringent assumption of identical dynamics, it is also limited since we assumed that collected demonstrations are optimal in the target environment, while it might not hold true in varying dynamics. In the future, we plan to address this combined challenge of multimodal OOD-IL and learning from sub-optimal demonstrations~\cite{suboptimal2019}.
}

\section*{Acknowledgments}
We would like to acknowledge the support of our wonderful coworkers: Yang Shu, Baixu Chen, Haixu Wu, and Yipeng Huang, whose generous help is an integral part of this work. 
This work was supported by the National Key Research and Development Plan (2020AAA0109201), National Natural Science Foundation of China (62022050 and 62021002), Beijing Nova Program (Z201100006820041), and BNRist Innovation Fund (BNR2021RC01002).
\bibliography{corl_2022}

\newpage

\appendix

\section{Contrastive Clustering Algorithm}

With the objectives introduced in the main text, we show our full contrastive clustering algorithm in Algorithm~\ref{alg:algo}. 

\begin{algorithm}[h]
\KwIn{Demonstrations $\Xi$, Feature extractor $F$, Cluster center Matrix $\mathbf{C}$, Sub-trajectory length $l$, Learning rate $\alpha$ and $\lambda$}
    Initialize the parameters of $F$.\\
    Randomly sample $K$ indices ${j_k}|_{k=1}^{K}$ from the interval $[1,N]$ \\
    Take a sub-trajectory $\xi^{\text{sub}}_{j_k}$  of length $l$ from $\xi_{j_k}$ and initialize $c_k$ with $F(\xi^{\text{sub}}_{j_k})$ \\
 \While{not converging}{
    Sample $N$ trajectories $\{\xi_n\}_{n=1}^{N}$ from $\Xi$ and subsample two sub-trajectories $\xi^{\text{sub}}_{2n-1}$ and $\xi^{\text{sub}}_{2n}$ of length $l$ for each $\xi_n \in \Xi$. \\
    Assign the cluster label $\mathbf{y}_{n}$ to $\xi^{\text{sub}}_n$ according to Eqn. (2).\\
    Update the parameters of $F$ by: $F \leftarrow F - \alpha \nabla_F \mathcal{L}_{\text{cluster}}$ according to Eqn. (3). \\
    Re-assign the cluster label $\mathbf{y}_{n}$ to $\xi^{\text{sub}}_n$ based on the updated $F$. \\
    Update the cluster centers $\mathbf{C}$ according to Eqn. (4). \
  }
  \For{$\xi \in \Xi$}{
  Take a uniformly sampled sub-trajectory $\xi^{\text{sub}}$ with length $l$ from $\xi$. \\
  Assign a cluster label $\mathbf{y}$ to $\xi$ according to Eqn. (2).\\ 
  }
  \KwOut{The cluster label $\mathbf{y}$ of each trajectory $\xi$.}
  \caption{Contrastive Clustering Algorithm}\label{alg:algo}
\end{algorithm}


\section{Details for Contrastive Clustering}
We further discuss the considerations of the design of contrastive clustering algorithm. Firstly, for varied-length sequences, it is difficult to design a proper distance metric to ensure that trajectories from the same mode are close, because common distance metrics such as per-step L2 or cosine distance on states cannot be used. Thus, contrastive learning is a good choice for learning the distance metric in a latent space for clustering. Secondly, separating contrastive learning and clustering into two stages may not find the optimal hidden space for clustering, while co-optimizing them can make them benefit from each other.

\textbf{Implementation Details}.
\noindent For the implementation of the contrastive learning algorithm, in the subsampling step, we fix the length of the sub-trajectories, which is no longer than $50$ steps since RNN usually suffers from catastrophic forgetting with long sequences. For the MuJoCo environment, the sub-trajectory length is fixed at $20$. For the Driving environment, the sub-trajectory length is fixed at $15$. For the Simulated Robot environment, the sub-trajectory length is fixed at $20$. For training, we randomly sample the sub-trajectories with a fixed stride and make sure each sub-trajectory has the same length. After convergence, we use the representation of a sub-trajectory for clustering. We set the batch size of contrastive clustering as $128$. We first pre-train the feature extractor only with the contrastive learning loss $\mathcal{L}_\text{contrast}$ for $200$ iterations before initializing $c_k$ and then train with the whole loss $\mathcal{L}_\text{cluster}$ for $2000$ iterations. For the number of clusters $K$, we set it as $5$, $10$, and $10$ at initialization for MuJoCo, Driving, and Robot Arm respectively. 
For the feature extractor, we use a one-layer LSTM model to extract representation for trajectories and set the dimension for the hidden state as $128$. The hyper-parameter $\lambda$ is fixed to $0.01$ for all three environments. Learning rate $\alpha$ is fixed to $0.01$ with Adam optimizer.

\section{Additional Experimental Results}
\subsection{Parameter Sensitivity}

\begin{figure}[htbp]
    \centering
     \subfigure[Expected return w.r.t. cluster number K]{\includegraphics[width=0.4\textwidth]{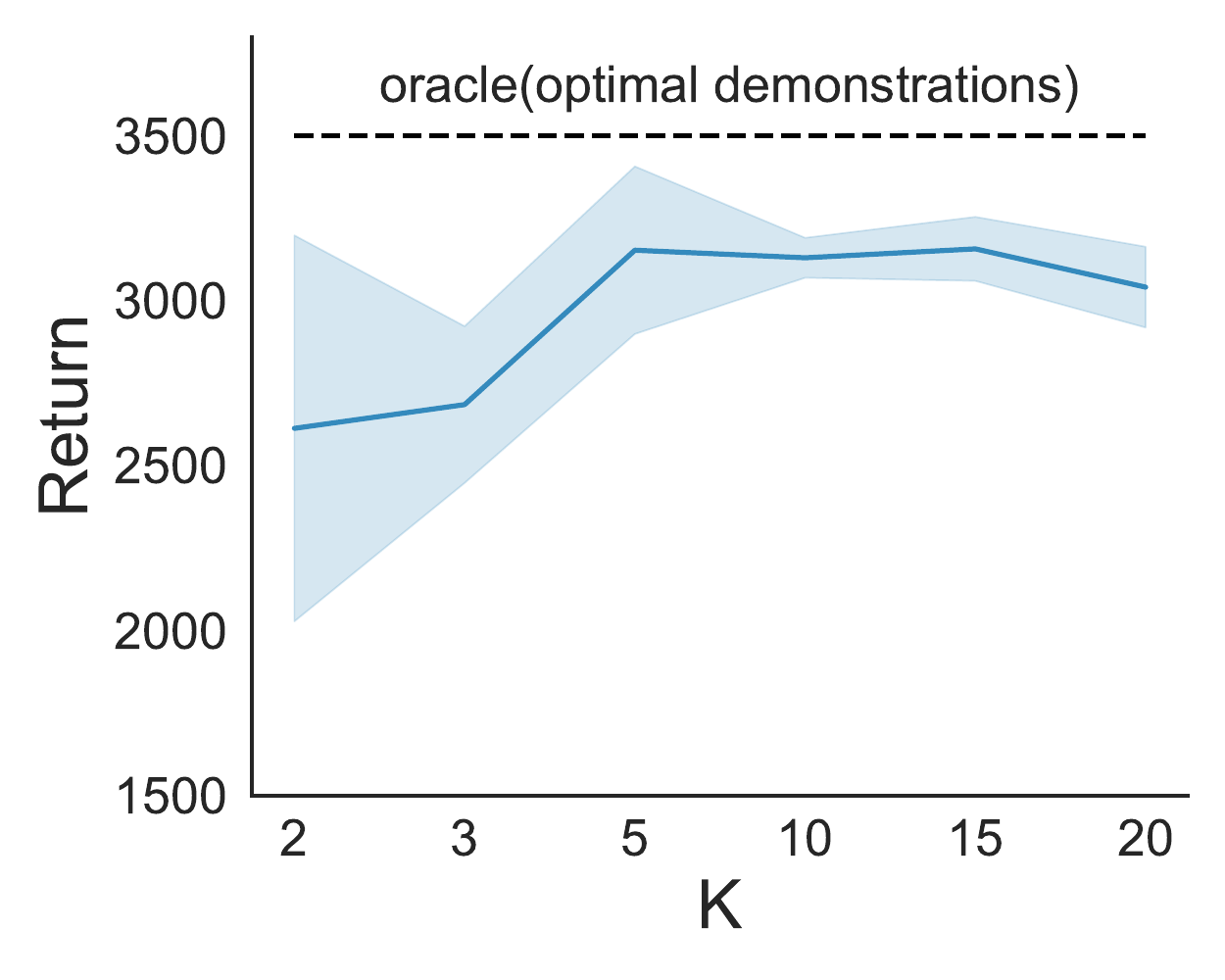}\label{fig:trendingK}}
      \subfigure[Expected return w.r.t. trade-off weight $\lambda$]{\includegraphics[width=0.4\textwidth]{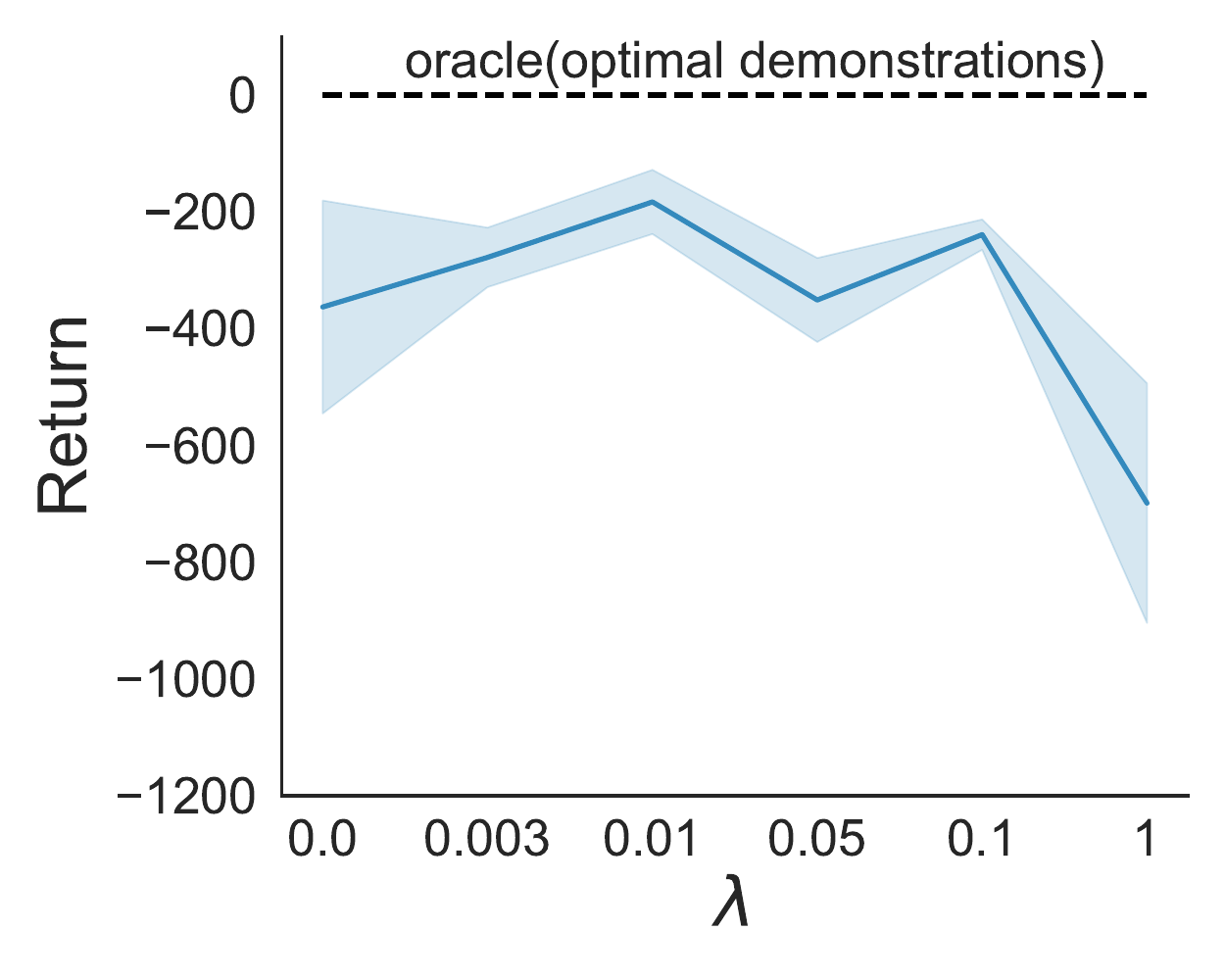}\label{fig:trendingBeta}}
     \caption{Sensitivity experiment results for hyperparameters $K$ and $\lambda$ }
     \label{fig:sensitivity}
\end{figure}

There are two key hyperparameters: cluster number $K$ and the trade-off weight $\lambda$ between $L_\text{contrast}$ and $L_\text{cluster} $ in our method. We investigate the sensitivity of the performance of our method to the hyperparameters.
We show the expected return after convergence under different hyperparameters in Figure~\ref{fig:sensitivity}. The solid lines with shades show the mean and standard deviation of the expected return of our method and the dashed lines on top show the oracle optimal performance that the policy may achieve by only selecting and learning from the optimal demonstrations.

 \textbf{Results.}  We observe that when $K$ is small, the converged model suffers from high variance and lower mean return, as a small number of clusters are not sufficient to capture all single modalities. Meanwhile, our method with larger $K$ achieves consistent performance, because more clusters guarantee a clear separation between different modalities. Once the cluster number is enough to capture all the modalities, more clusters do not improve the performance. Nevertheless, we note that larger $K$ brings extra computational cost since every cluster requires training a GAIL model, so we set $K$ to $5$, $10$, and $10$ respectively for $3$ environments in consideration of the trade-off between efficiency and effectiveness.  For the sensitivity of $\lambda$, we find that our framework works well under the value of $\lambda$ ranging from $0.003$ to $0.1$, and $\lambda > 1$ leads to a severe drop in performance, mainly because too much emphasis on $L_\text{cluster}$ will cause all samples to collapse into one or two clusters and the contrastive clustering algorithm becomes unable to separate different modalities.
 
\textbf{Discussion on the choice of $K$.}
\textcolor{revisioncolor}{While in real scenarios when $K$ is unknown to us, we can estimate it empirically by dimension reduction and then visualizing trajectories. If one wants to get an optimal K, a grid search around this approximation may be needed, but often an approximation is good enough. Note that the number of modes has no direct relationship with the number of source domains, especially in a real-world scenario: data can be collected every day, and each day can be seen as a source, but they may all fall into a certain number of modes, i.e., the number of modes will not increase unlimitedly. After contrastive clustering, 
transferability learning on each cluster can be done in parallel, which can save time. Only a subset of demonstrations can also be easier to fit, compared to fitting the whole dataset, which also boosts learning efficiency.  }

\subsection{Visualization of Transferability}
\begin{figure}[htbp]
        \centering
        \includegraphics[width=0.95\textwidth]{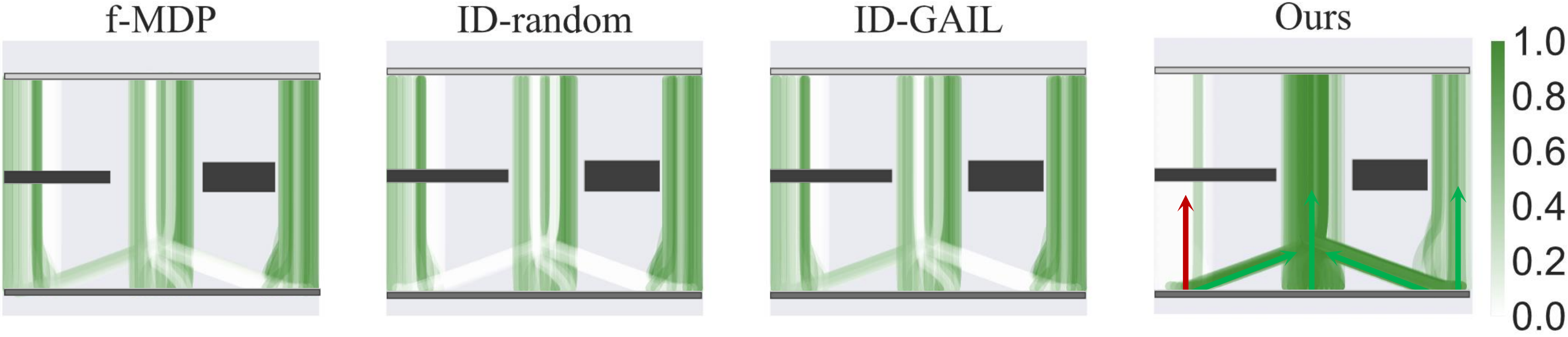}
        \caption{Visualization of the transferability in Driving for different methods.}
        \label{fig:visualize}
    \end{figure}

We visualize the transferability computed by the proposed method as well as all our baselines on the Driving environment. As is shown in Figure~\ref{fig:visualize}, the deeper the color, the higher transferability of the trajectory. We can observe that our method can mostly filter out non-transferable demonstrations (red arrow) for the target environment while assigning high transferability for transferable ones (green arrow).    
 
 \vspace{-10pt}

 \subsection{Effect of the Ratio of Transferable and Non-transferable Demonstrations}
 To investigate the influence of the composition of the demonstrations on the final imitation performance, we conduct experiments on \textcolor{revisioncolor}{three MuJoCo environments} with different ratios of the demonstrations from the four source demonstrators. For Hopper, we set the gravitational constant as (i) $15.0$, (ii) $9.8$, (iii) $2.0$, (iv) $1.0$. We fix the number of trajectories for (i) and (ii), i.e. relatively transferable demonstrations, and change the number of trajectories for (iii) and (iv). 
 \textcolor{revisioncolor}{For Walker2d, we set friction to (i) $24.8$, (ii) $9.9$, (iii) $3.9$, (iv) $1.1$. } \textcolor{revisioncolor}{For HalfCheetah, the compositions of demonstrations are set the same as in original paper, which is (i) ($1$, $0.9$), (ii) ($0.9$, 1), (iii) ($1$, $0.05$), (iv) ($0.05$, $1$) with setting ($\cdot$, $\cdot$) as the discount factor of the force of the front leg and the back leg.}
 The ratio configurations and the results are shown in \textcolor{revisioncolor}{Table~\ref{tab:diffratio}, Table~\ref{tab:diffratioWalker}, Table~\ref{tab:diffratioHalfCheetah}  respectively for three environments.}
 
 We observe that in the Hopper environment, with the increase of non-transferable trajectories, the performance of naive GAIL deteriorates and other baselines also drop dramatically due to the multimodal distribution effect, while our method shows stable performance with a high return against the changes in the composition of the source demonstrations. Moreover, comparing with the converged result of our method under different ratios, we observe that increasing non-transferable trajectories does not influence the final return of our method much, which indicates that our transferability measurement stably and accurately filters out non-transferable trajectories. Even for the easiest setting: $1 : 1 : 1: 1 $ with an equal number of transferable and non-transferable demonstrations, our method still outperforms GAIL. The results show that non-transferable demonstrations consistently influence imitation learning performance and measurement to filter out non-transferable demonstrations is important. 
\textcolor{revisioncolor} {Experiments on the other environments show similar results, demonstrating the robustness of our method under various scenarios.}
 
 \begin{table}[htbp]
        \centering
        \renewcommand{\arraystretch}{1.1}
        \caption{Results under different compositions of demonstrations in Hopper environment.}
                \vspace{5pt}
            \setlength{\tabcolsep}{1.6mm}{
                \begin{tabular}{c|ccccc}
                    \toprule

                    {Composition } 
                    & Naive GAIL  &   fMDP &  ID w/o GAIL  &   ID w/ GAIL  & Ours \\  \midrule
                    $1 : 1 : 1: 1 $   & 2926$\pm$468 & 2947$\pm$412 & 1547$\pm$362 & 2287$\pm$315 & \textbf{3259}$\pm$198 \\ 
                    $1 : 1 : 2: 2 $   & 2845$\pm$360 & 2662$\pm$699 & 1335$\pm$787 & 2022$\pm$253 &  \textbf{3261}$\pm$206  \\ 
                    $1 : 1 : 5: 5 $   & 2761$\pm$358 & 2361$\pm$537 & 1176$\pm$154 &  1042$\pm$730 &  \textbf{3104}$\pm$340 \\ 
                    $1 : 1 : 10: 10 $   & 2137$\pm$685  & 2791$\pm$468 & 836$\pm$218 & 1314$\pm$412 &  \textbf{3049}$\pm$331 \\ 
                    $1 : 1 : 25: 25 $   & 1083$\pm$244  & 1040$\pm$760 & 908$\pm$191 & 714$\pm$82 &  \textbf{3113}$\pm$413  \\ 
                    $1 : 1 : 50: 50 $   & 739$\pm$184  & 1276$\pm$458 & 764$\pm$260 &  671$\pm$126 &  \textbf{2890}$\pm$556 \\     \bottomrule
                    
                \end{tabular}
                }
            
            \label{tab:diffratio}
        \end{table}
        \begin{table}[htbp]
            \centering
            \renewcommand{\arraystretch}{1.1}
            \caption{\textcolor{revisioncolor}{Results under different composition of demonstrations in Walker2d environment.}}
                \vspace{5pt}
                \setlength{\tabcolsep}{1.6mm}{
                    \begin{tabular}{c|ccccc}
                        \toprule

                        {Composition } 
                        & Naive GAIL  &   fMDP &  ID w/o GAIL  &   ID w/ GAIL  & Ours \\  \midrule
                        $1:2:2:2  $            & 318$\pm$290            & 283$\pm$190           & 1688$\pm$218          & 1703$\pm$175          & \textbf{2077}$\pm$216          \\
                        $1:2:5:5    $          & 288$\pm$72             & 249$\pm$37            & 345$\pm$92            & 328$\pm$39            & \textbf{1731}$\pm$73           \\
                        $1:2:10:10    $        & 327$\pm$65             & 213$\pm$48            & 311$\pm$29            & 349$\pm$30            & \textbf{1664}$\pm$166           \\
                        $1:2:20:20   $         & 287$\pm$74             & 339$\pm$131           & 345$\pm$73            & 320$\pm$64            & \textbf{1629}$\pm$87           \\
                             \bottomrule
                        
                    \end{tabular}
                    }
                
                \label{tab:diffratioWalker}
            \end{table}
        \begin{table}[htbp]
            \centering
            \renewcommand{\arraystretch}{1.1}
            \caption{\textcolor{revisioncolor}{Results under different composition of demonstrations in HalfCheetah environment.}}
                \vspace{5pt}
                \setlength{\tabcolsep}{1.6mm}{
                    \begin{tabular}{c|ccccc}
                        \toprule

                        {Composition } 
                        & Naive GAIL  &   fMDP &  ID w/o GAIL  &   ID w/ GAIL  & Ours \\  \midrule
                        $2:2:1:1  $            & 2389$\pm$897            & 404$\pm$246           & 2031$\pm$312          & 2210$\pm$86          & \textbf{3008}$\pm$117          \\
                        $2:2:2:2    $          & 2882$\pm$84             & 247$\pm$308            & 2126$\pm$110            & 2067$\pm$86            & \textbf{2997}$\pm$209           \\
                        $2:2:5:5    $        & 2201$\pm$502             & 1613$\pm$409            & -327$\pm$119            & 1273$\pm$546            & \textbf{3246}$\pm$134           \\
                        $2:2:10:10   $         & 2367$\pm$897             & 389$\pm$232           & 1808$\pm$146            & 1315$\pm$414            & \textbf{2981}$\pm$71           \\
                             \bottomrule
                        
                    \end{tabular}
                    }
                
                \label{tab:diffratioHalfCheetah}
            \end{table}

\subsection{Additional Experiments on Simulated Robot}

\begin{figure}[h]
    \centering
    \subfigure[]{\includegraphics[width=0.46\textwidth]{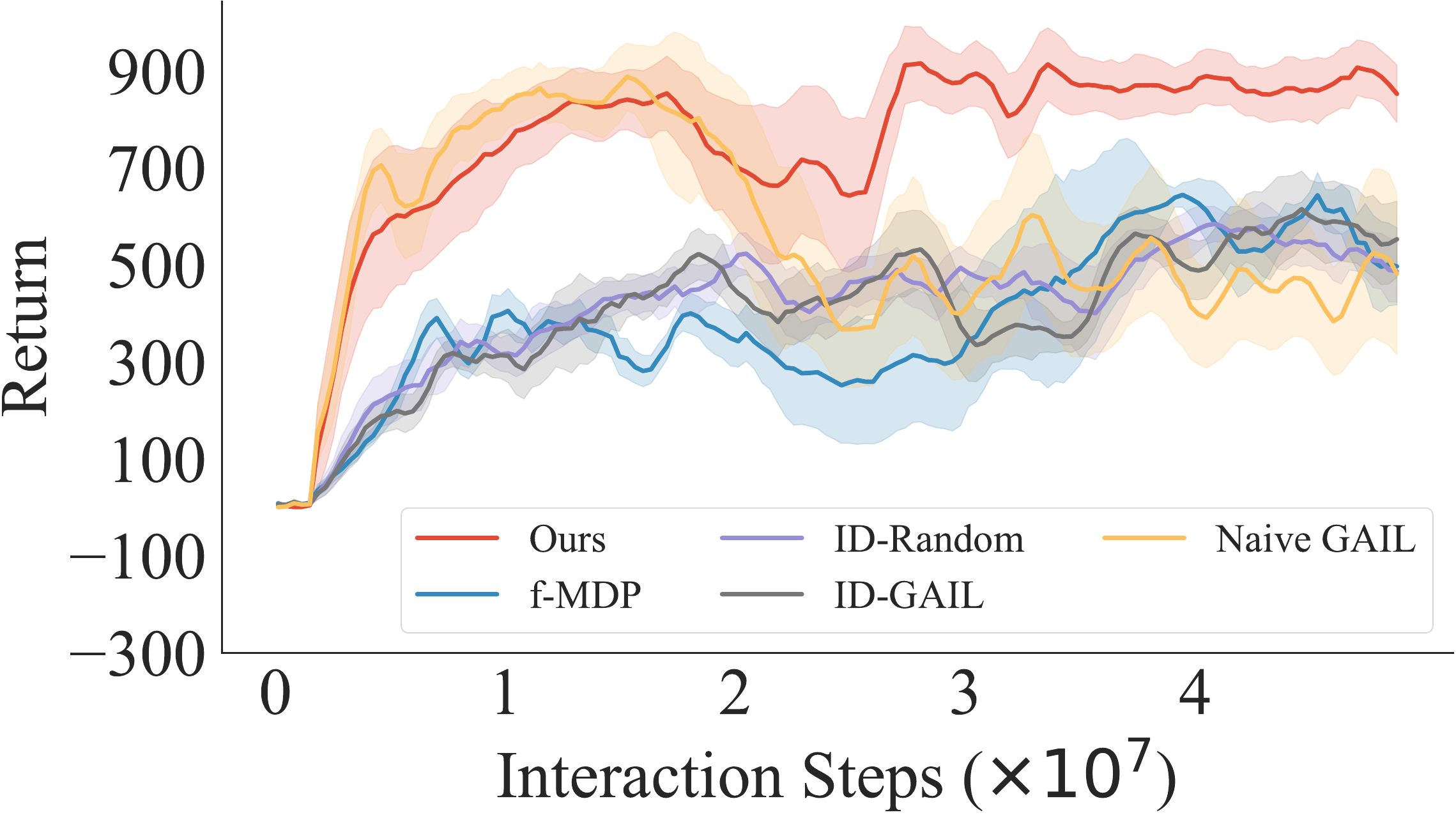}\label{fig:result13}}
    \subfigure[]{\includegraphics[width=0.5\textwidth]{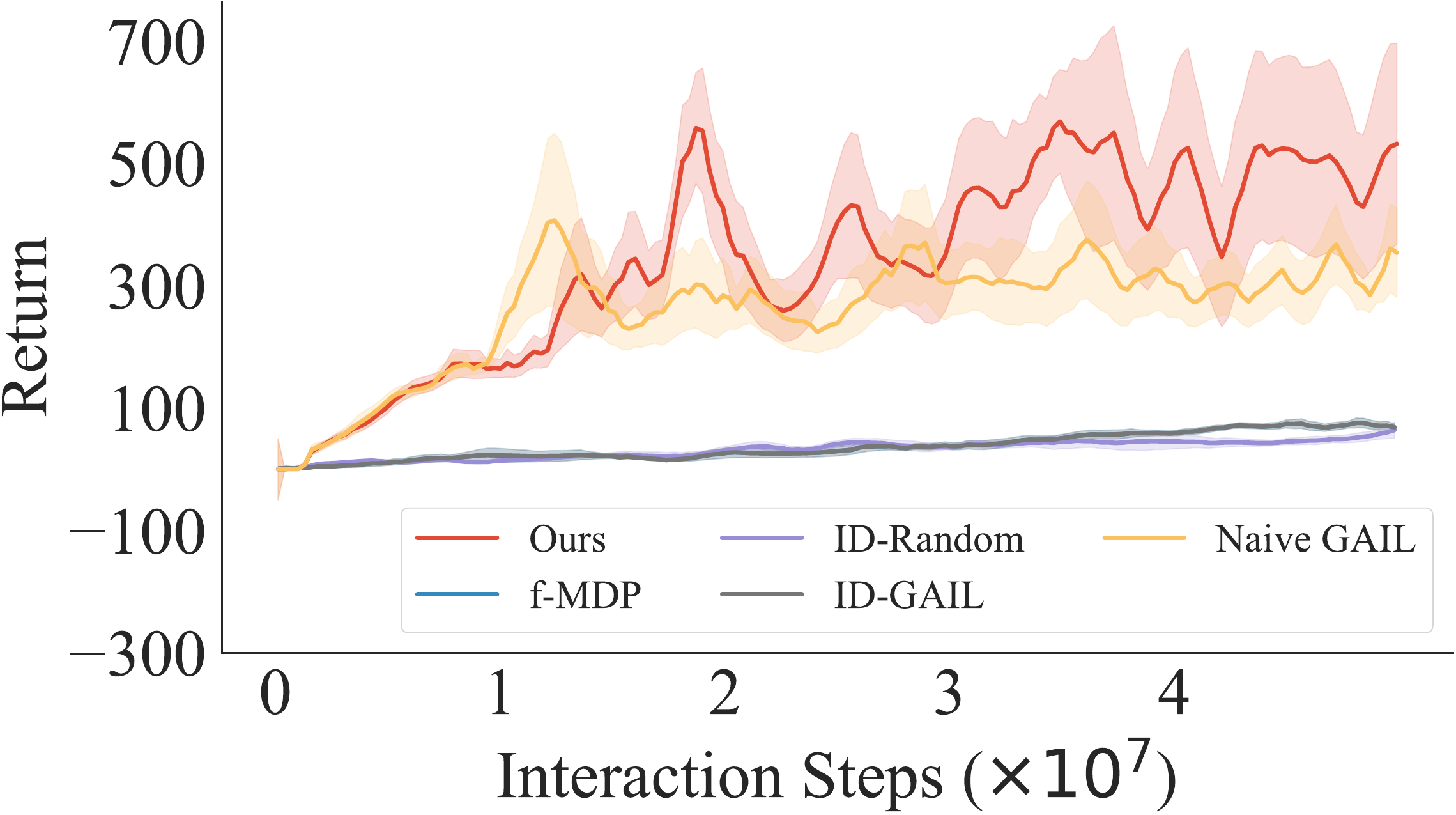}\label{fig:result134}}
    \caption{Additional experiment results on simulated robot environment.}
\end{figure}

We also conduct additional experiments on simulated Franka Panda Arm to better verify our proposed method. We create three demonstrators by disabling the No. $1,3$ joints, the No. $1$ joint, and using fully-able joints respectively while disabling the No. $1, 3$ joints for the target imitator. We import  demonstrations with the number of interaction steps  $1\times 10^5$, $1\times 10^5$, and $1\times 10^5$ for each source environment respectively. The reward function and the task are set as the same as that of the original task in the main paper. The result is shown in Fig.~\ref{fig:result13}. Another setting is created similarly by disabling the No. $1,3,4,6$ joints, the No. $1,3$, and the No. $4$ joint respectively while disabling the No. $1, 3,4$ joints for the target imitator, with the result presented in Fig.~\ref{fig:result134}. 

\subsection{Generalization to More Demonstrations} In real-world applications, there are situations where demonstrations in the original database are insufficient and new demonstrations are continuously collected from different sources to augment the database. We further demonstrate that the proposed method can use augmented demonstrations more effectively. We conduct experiments in the MuJoCo Walker2d experiment. We firstly collect $2$, $2$, $50$ and $50$ demonstrations from environment (i) $24.8$, (ii) $9.9$, (iii) $3.9$, (iv) $1.1$ respectively. The demonstrations are not enough to learn an optimal policy, but our method can still learn a transferability model and f-MDP and ID can learn a feasibility model. Then we add $10$, $50$, and $50$ demonstrations from environment (v) $24.9$, (vi) $0.7$, and (vii) $0.1$ respectively. Then we require all the methods not to re-train the transferability or the feasibility model but directly predict the transferability or feasibility for new demonstrations. The experiments aim to test the generalization ability of the model to filter out non-transferable demonstrations. 

\begin{figure}[htbp]
    \centering
    \subfigure[Results on insufficient trajectories. ]{\includegraphics[width=0.48\textwidth]{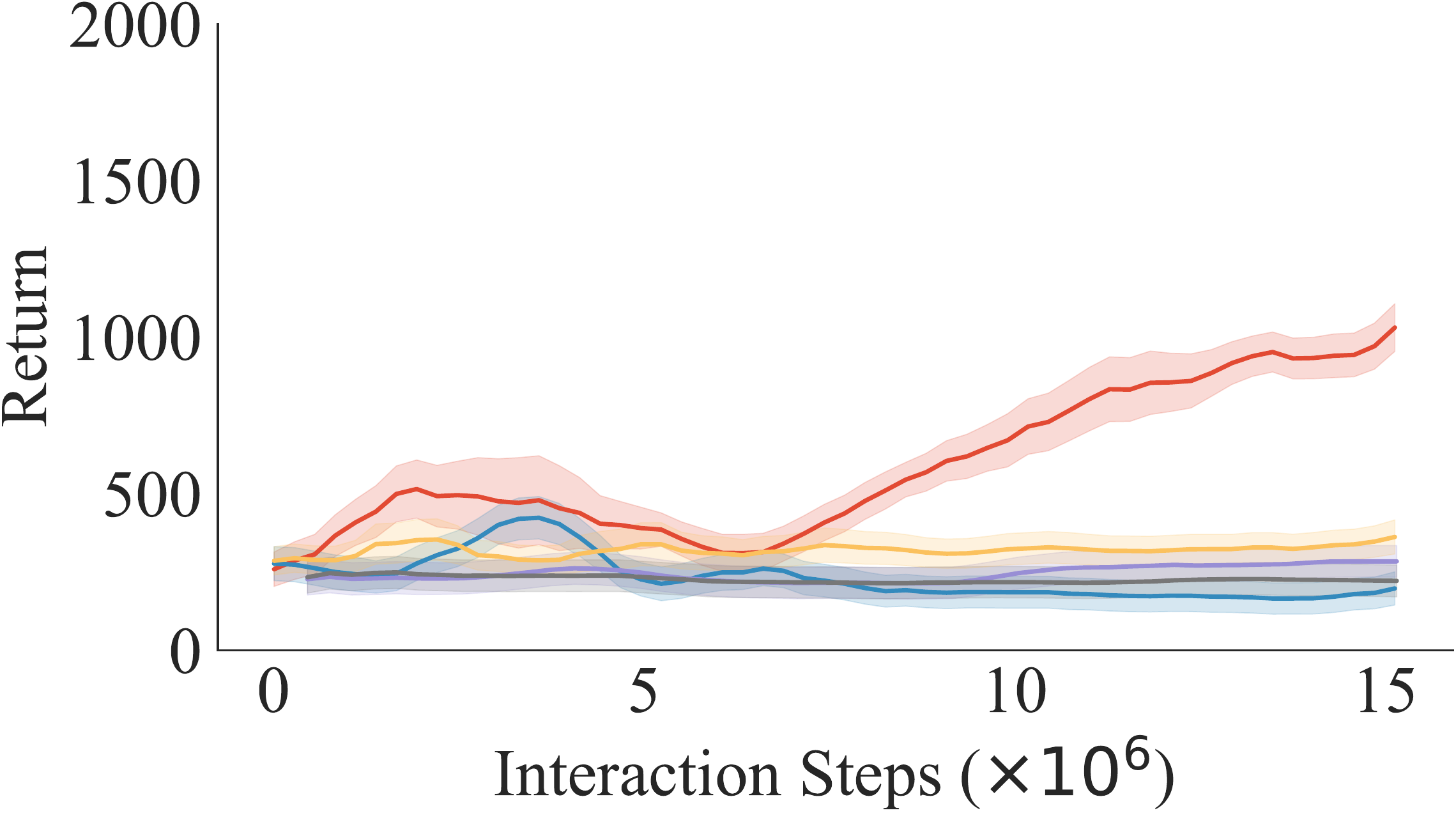}\label{fig:generalize1}}
     \subfigure[Generalization to newly given demonstrations.]{\includegraphics[width=0.48\textwidth]{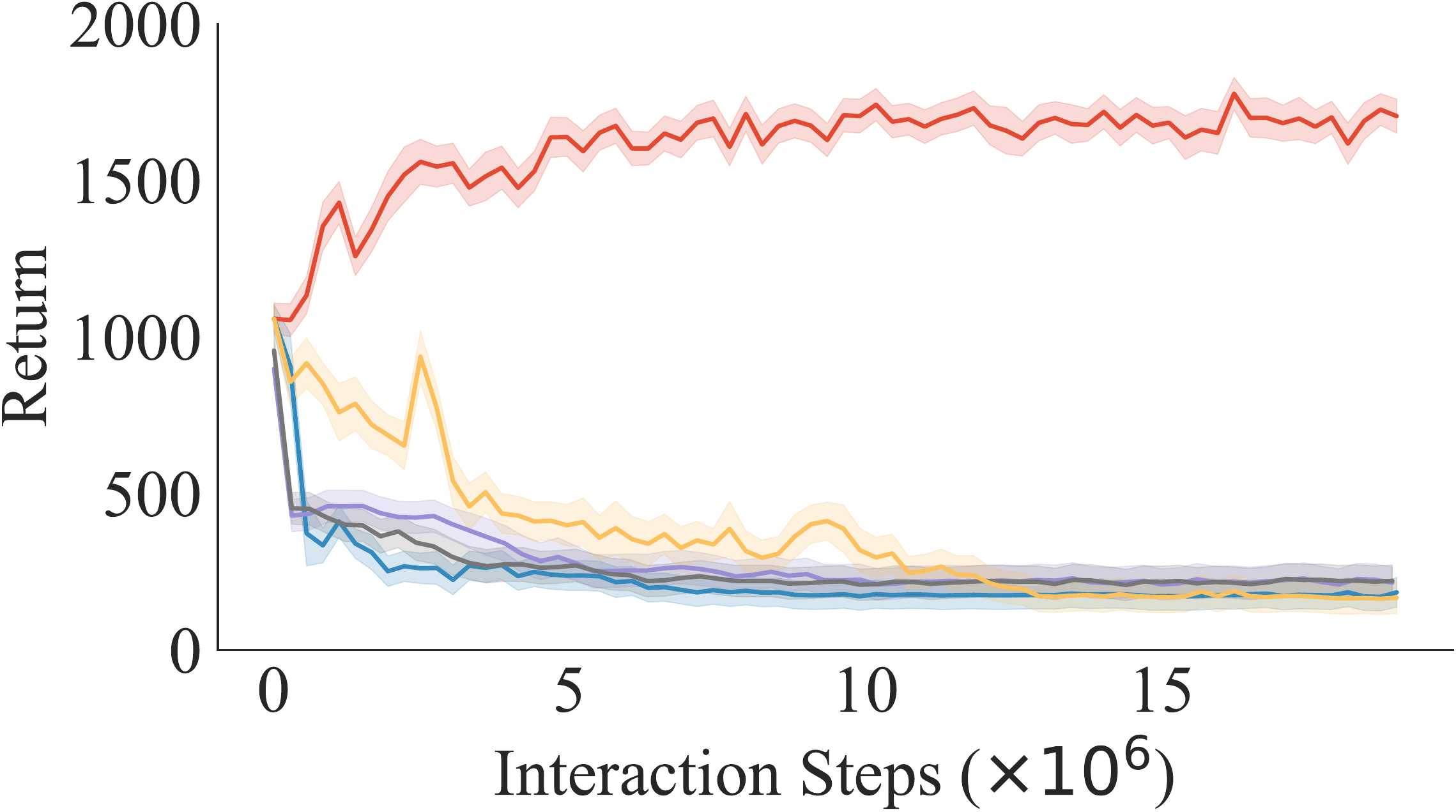}\label{fig:generalize2}}
    \caption{Experiments for generalization to more demonstrations. }
    \label{fig:generalize}
\end{figure}

As shown in Fig.~\ref{fig:generalize1},  when we only have insufficient demonstrations, we observe that the proposed method still achieves the highest point compared with other baselines, which demonstrates that we are able to use the demonstrations more efficiently even when they are insufficient.

Moreover, in Fig.~\ref{fig:generalize2}, we are given new demonstrations. To use them selectively, we first use our contrastive-clustering LSTM model to assign a cluster label to each demonstration according to Eqn. (2). We then generate the transferability for the new demonstrations with the GAIL model in that cluster according to Eqn. (6). Note that we do not re-train the clustering model here with the new demonstrations but directly apply the clustering model and the GAIL model for transferability to cluster new demonstrations. For a fair comparison, we finetune the policy starting from the same checkpoint achieved by our method. The proposed method achieves the highest performance, which means that the proposed method possesses the capability of generalizing to unseen demonstrations. This generalization to new demonstrations can be extremely meaningful, which serves as a practical method to satisfy our intention of \textbf{continually} collecting more useful information from multiple sources. We do not require any extra computation other than a one-time inference, which is efficient to use.
\begin{figure}[htbp]
    \centering
    \includegraphics[width=.8\textwidth]{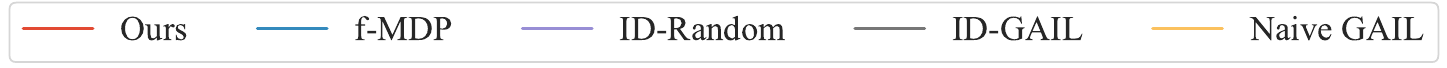}
\end{figure}
\vspace{-10pt}
\vspace{10pt}

\subsection{Comparison with a K-means Variant}

\begin{wrapfigure}{r}{0.5\textwidth}
    \centering
     \includegraphics[width=0.45\textwidth]{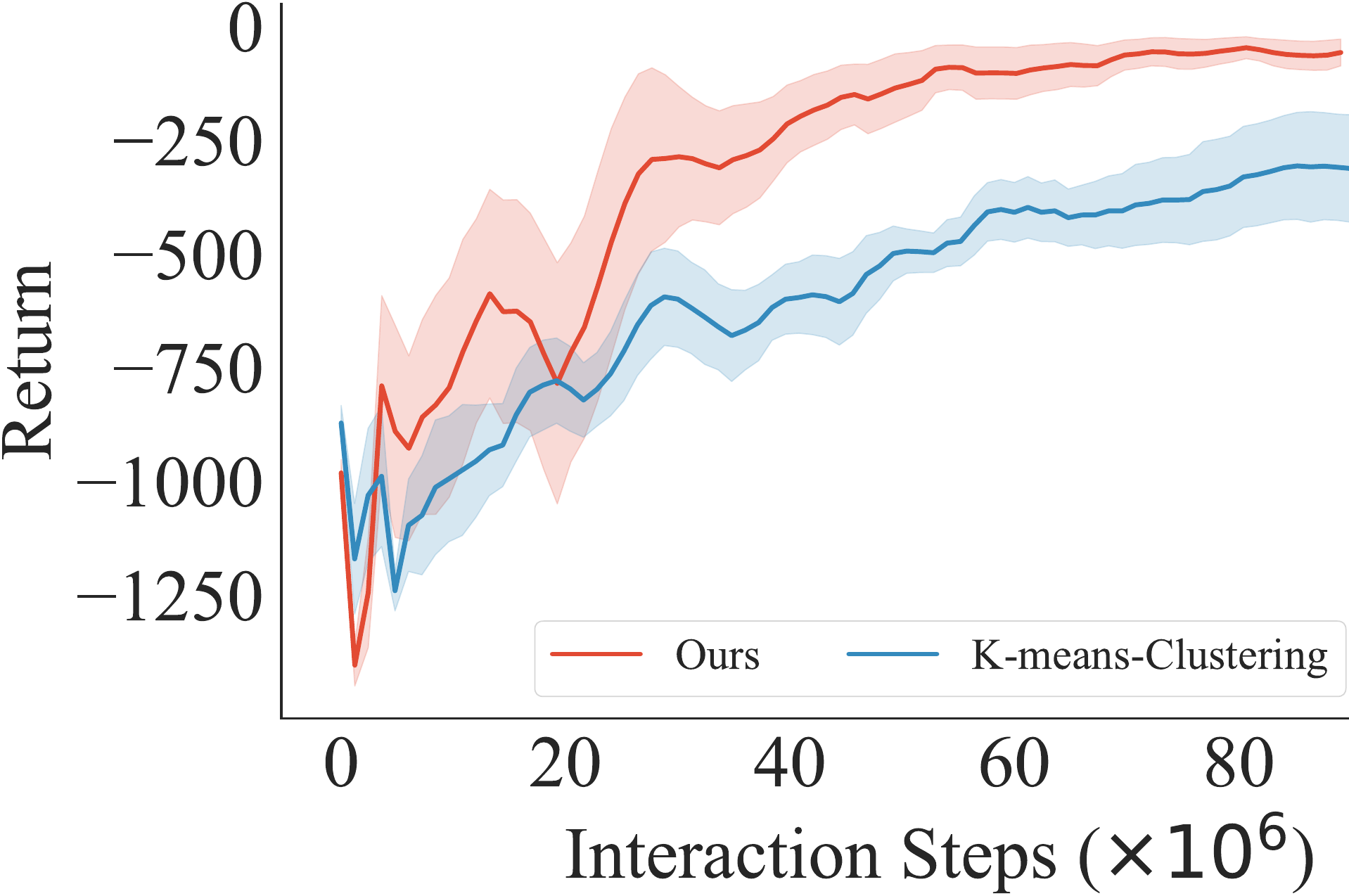}
    \caption{The ablation study on a K-means variant of our method.}
    \label{fig:ablation-Kmeans}
    \vspace{-10pt}
\end{wrapfigure}

To demonstrate the significance of our Sequence-based Contrastive Clustering algorithm, we conducted the following experiments on the Driving environment by K-Means clustering with the number of clusters $K$=$10$ (as the same in our method).

Specifically, we down-sampled each trajectory with a fixed stride to uniformly generate a fixed-length subsample, and applied the K-means algorithm directly to these sub-trajectories and therefore assign each trajectory to a cluster. Then, on each of these $K$ clusters, we learn the transferability respectively. The result of using transferability generated by K-Means clustering for the final imitation learning is presented in Fig.~\ref{fig:ablation-Kmeans}. 

We observed that the lacking of a contrastive learning step may cause difficulty in obtaining a high-quality unimodal clustering, which is essential for learning an accurate transferability measurement, and further cause a final performance drop. One way our contrastive clustering method is superior to K-means is that performing K-means on uniformly random-sampled sub-trajectories may introduce high variance into the clustering results, while our method, which makes different subsamples of the same trajectory as positive pairs and minimize their distance in the hidden representation space, can mitigate such instability. Also, the extracted representations are used for clustering, so it is beneficial if they are learned with the clustering step in a coherent manner.


\end{document}